%% file: cvpr.tex
\documentclass[final]{latex/cvpr}

\usepackage{times}
\usepackage{epsfig}
\usepackage{graphicx}
\usepackage{amsmath}
\usepackage{bm}

\usepackage[noend]{algpseudocode}
\usepackage{algorithmicx,algorithm}

\usepackage[usenames, dvipsnames]{color}

\usepackage[switch]{lineno}
\usepackage{lineno}
\usepackage{booktabs}
\usepackage{multirow}
\usepackage{appendix}
\usepackage{balance}

\usepackage{setspace}

\usepackage[pagebackref=true,breaklinks=true,colorlinks,bookmarks=false]{hyperref}
\def\cvprPaperID{679} 
\def\confYear{CVPR 2021}

\begin{document}

\title{QAIR: Practical Query-efficient Black-Box Attacks for Image Retrieval}



\author{Xiaodan Li\textsuperscript{1,}\textsuperscript{$\ast$}, Jinfeng Li\textsuperscript{1}, Yuefeng Chen\textsuperscript{1}, Shaokai Ye\textsuperscript{2}, Yuan He\textsuperscript{1}, Shuhui Wang\textsuperscript{3}, Hang Su\textsuperscript{4}, Hui Xue\textsuperscript{1}\\
\textsuperscript{1}Alibaba Group \quad\quad \textsuperscript{2}EPFL \quad\quad \textsuperscript{3}Inst. of Comput. Tech., CAS, China\\ \textsuperscript{4} Institute for AI, THBI Lab, Tsinghua University, Beijing, 100084, China\\
\tt\small \{fiona.lxd, jinfengli.ljf, yuefeng.chenyf, heyuan.hy, hui.xueh\}@alibaba-inc.com\\
\tt\small shaokaiyeah@gmail.com, wangshuhui@ict.ac.cn, suhangss@mail.tsinghua.edu.cn
}

\maketitle




\renewcommand{\thefootnote}{\fnsymbol{footnote}}
\footnotetext[1]{Corresponding author}

\begin{abstract}
We study the query-based attack against image retrieval to evaluate its robustness against adversarial examples under the black-box setting, where the adversary only has query access to the top-$k$ ranked unlabeled images from the database.
Compared with query attacks in image classification, which produce adversaries according to the returned labels or confidence score, the challenge becomes even more prominent due to the difficulty in quantifying the attack effectiveness on the partial retrieved list.
In this paper, we make the first attempt in Query-based Attack against Image Retrieval~(QAIR), to completely subvert the top-$k$ retrieval results.
Specifically, a new relevance-based loss is designed to quantify the attack effects by measuring the set similarity on the top-$k$ retrieval results before and after attacks and guide the gradient optimization.
To further boost the attack efficiency, a recursive model stealing method is proposed to acquire transferable priors on the target model and generate the prior-guided gradients.
Comprehensive experiments show that the proposed attack achieves a high attack success rate with few queries against the image retrieval systems under the black-box setting.
The attack evaluations on the real-world visual search engine show that it successfully deceives a commercial system such as Bing Visual Search with 98\% attack success rate by only 33 queries on average.
\end{abstract}

\section{Introduction}
\input{latex/intro_4}

\section{Related Work}
\label{sec:query-based}
\input{latex/related_2}

\section{Methodology}
\input{latex/method}

\section{Experiments}

\input{latex/experiment}

\section{Conclusion}
\input{latex/conclusion}

\newpage

{\small
\bibliographystyle{latex/ieee_fullname}
\bibliography{latex/egbib}
}

\include{latex/appendix}

\end{document}

%% file: latex/intro_4.tex
\begin{figure}[t]        
  \center{\includegraphics[width=0.8\linewidth]{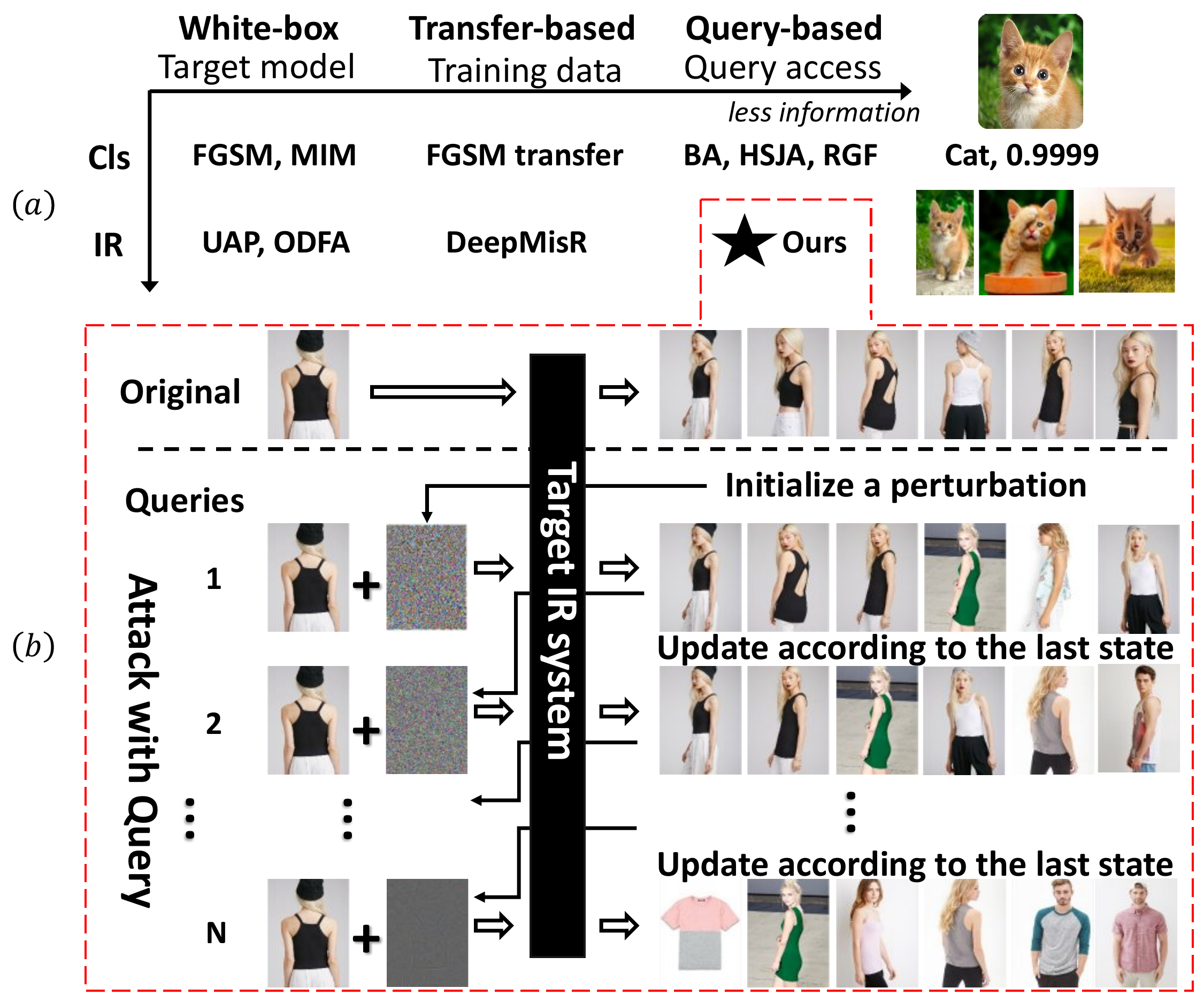}}
 \caption{\label{fig:diff_cls_ret_2} 
 (a) (Left)Taxonomy of adversarial attacks. Different from existing attacks for IR, our attack is applicable to real-world scenarios since it only needs query access to the target model. (Right) The output of image classification(Cls) is a label or confidence score while it is a list of unlabeled images in IR.\\
 (b) Demonstration of the query-based black-box attack on IR. Given a target model, the attackers use queries to update and generate adversarial perturbations.}
 \end{figure}
\vspace{-3mm}

Despite of its impressive performance in many tasks such as image classification~\cite{he2016deep}, object detection~\cite{chen2018shapeshifter} and image retrieval (IR)~\cite{wang2019multi}, 
deep neural network~(DNN) has been shown to be vulnerable to adversarial examples that can trigger the misbehavior with human-imperceptible perturbations~\cite{goodfellow2014explaining, cheng2019improving,DPQN}.
Such vulnerability has raised great concerns about the robustness and real-world deployment of DNNs for image retrieval~\cite{li2019universal,zhao2019unsupervised} and object detection~\cite{chen2018shapeshifter}, {\it et al.}.
For example, in digital right management, the original graphic designs are protected by checking if there exists a same design in the top-$k$ similar ones retrieved from the whole graphic design database. 
By adding adversarial perturbations on protected designs, attackers can deceive the target IR system into retrieving some irrelevant images
for evading the censorship of professional monitors.
Therefore, it is crucial to develop a practical robustness evaluation technology to explore the vulnerability of IR systems against adversarial attacks, and then facilitate the development of the corresponding countermeasures.

A general idea of adversarial attack is to generate adversarial examples with human-imperceptible perturbations along the gradient direction by {\it maximizing} a certain loss function, {\it e.g.}, a classification loss~\cite{goodfellow2014explaining, dong2020benchmarking}.
However, as shown in Fig.~\ref{fig:diff_cls_ret_2}, the IR system produces a list of images for a query input.
This makes it hard to define the objective function for indicating the attack effectiveness only with the retrieved list. Under this circumstance, the gradients can hardly be estimated for deriving effective attacks.
Though there exist some decision-based methods~\cite{brendel2017decision, chen2020hopskipjumpattack} in attacking classification models which only rely on the final decision to indicate whether the attack succeeds, they usually require a significant attack cost by a tremendous number of queries to cross the decision boundary via greedy search~\cite{cheng2019improving}.

Furthermore, in adversarial attacks, the gradients for guiding the attack process are usually calculated based on the knowledge of the target model, {\it e.g.}, the model structure and parameters.
Existing studies on adversarial attacks against IR systems mainly focus on the white-box attacks in which attackers are assumed to have complete knowledge about the target model~\cite{zhou2020adversarial, li2019universal,wang2020transferable,shan2020fawkes}, so the gradients can be directly acquired.
However, the underlying white-box assumption does not hold in reality. 
Some studies try to use an approximate gradient instead for crafting adversarial examples~\cite{nesterov2017random,wang2020transferable}. The approximate gradient could be either the gradient of a surrogate model ({\it a.k.a.} transfer-based attacks) or numerically estimated by methods ({\it a.k.a.} query-based attacks) such as the zero-order optimization~\cite{chen2017zoo}.
The transfer-based methods attack the target model by leveraging adversarial examples generated against a white-box substitute model~\cite{wang2020transferable, li2019universal}, requiring training data that are usually protected. 
Besides, their attack success rate is still unsatisfactory due to the lack of adaptation procedure when the generated adversarial examples fail to attack the target model~\cite{cheng2019improving}.
The query-based attacks produce the gradient with methods such as finite difference~\cite{chen2017zoo, bhagoji2018practical}, random gradient estimation~\cite{nesterov2017random}. However, they are not efficient enough due to the lack of knowledge about the target model.

To address the aforementioned challenges, we propose the first attempt on practical Query-efficient Attack against Image Retrieval (QAIR) under the black-box setting.
First of all, we formulate the problem of black-box attacks on IR systems, and propose a new relevance-based loss to quantify the attack effects on target models with probabilistic interpretation. 
In this way, the structural output of IR systems can help to guide the gradient estimation during attacks.
Besides, considering the fact that retrieved images are ranked based on similarities with the input image, which can generate plenty of labeled triplets, a recursive model stealing method is constructed on the ranking list to acquire transfer-based priors and generate the prior-guided gradients.
Extensive experiments show that the proposed method can achieve a high attack success rate against IR systems with a remarkable Recall@$K$ drop. We also evaluate our attack efficacy on the real visual search system\footnote{https://www.bing.com/visualsearch\label{bing}}, which demonstrates its practicability in real-world scenarios.


Our main contributions can be summarized as follows:
\begin{itemize}
\item We formulate the problem of black-box attacks against image retrieval systems, and propose a new relevance-based loss to quantify the attack effects.
\item We develop a recursive model stealing method to acquire transfer-based priors of target model for boosting the query-attack efficiency.
\item We demonstrate the efficacy of our attack through extensive experiments on simulated environments and real-world commercial systems.
\end{itemize}


%% file: latex/related_2.tex
In this section, we briefly introduce image retrieval and review existing adversarial attacks.

\subsection{Image Retrieval}
Image retrieval is a popular topic in computer vision and has been widely used in commercial systems such as Google Image Searching\footnote{https://images.google.com/\label{google}}, Bing Visual Search\textsuperscript{\ref{bing}}, {\it etc.}~\cite{li2019universal}. 
A deep metric learning based image retrieval system usually consists of a metric learning model ({\it a.k.a} image retrieval model) and a database (known as gallery)~\cite{wang2019multi}.
Given a query image, the metric learning model will extract and compare its feature with images in the gallery, then retrieve related ones based on their similarities with the query.

The metric learning model can be different in terms of training strategies.
For example, contrastive loss~\cite{hadsell2006dimensionality} is proposed to make representations of samples from positive pairs to be closer while those from negative pairs to be far apart.
Some researchers claim that pair-wise metric learning often generates a large amount of pair-wise samples, which are highly redundant. Training with random sampling may significantly degrade the model capability and also slow the convergence.
Thus, hard mining strategy~\cite{schroff2015facenet} and lifted structure loss~\cite{oh2016deep} are proposed.
Recently, multi-similarity loss~\cite{wang2019multi} is proposed to establish a general pair weighting framework to formulate deep metric learning into a unified view of pair weighting and has achieved a state-of-the-art performance.

\subsection{Adversarial Attack}
Adversarial examples are maliciously crafted by adding human-imperceptible perturbations that trigger DNNs to misbehave~\cite{goodfellow2014explaining}. 
The attacks for generating adversarial examples can be summarized into white-box~\cite{goodfellow2014explaining,madry2017towards}, transfer-based ~\cite{dong2020benchmarking} and query-based attacks~\cite{brendel2017decision,chen2020hopskipjumpattack} in terms of the information that attackers rely on~\cite{dong2020benchmarking}. The gradient calculation also differs a lot among these kinds of attacks.

\textbf{White-box.}
Under the white-box setting, attackers have full access to the target model and they can directly acquire the true gradient of the loss 
{\it w.r.t.} the input. For instance, Opposite Direction Feature Attack~(ODFA)~\cite{zheng2018open} generates adversarial examples by querying the target model's parameters and pushing away the feature of adversarial query in the opposite direction of their initial counterparts. 
To generate image-agnostic universal adversarial perturbations~(UAP), Li~{\it et al.}~\cite{li2019universal} try to optimize the traditional triplet loss inversely against metric learning on feature embeddings. However, the underlying white-box assumption usually does not hold in real-world scenarios.

\textbf{Transfer-based.} 
Transfer-based attacks do not rely on model information but need information about the training data to train a fully observable substitute model~\cite{papernot2017practical, guo2019subspace}.
For instance, DeepMisRanking (DeepMisR)~\cite{bai2019metric} deceives the target models based on the transferability of adversarial examples generated against the substitute model by a white-box attack.
But the training data may be unavailable in real applications. Though some work~\cite{zhou2020dast, kariyappa2020maze} propose to steal model in a data-free manner, {\it e.g.}, producing inputs by generative models, this issue needs to be further investigated in image retrieval tasks. Besides, the performance of transfer-based attacks is limited due to the lack of adjustment when the gradient of the surrogate model points to a non-adversarial region of the target model~\cite{cheng2019improving}.

\textbf{Query-based.}
The query-based attack is more practical since the adversary in reality usually only has query access to the output of the target model.
This kind of attack has been widely studied in the task of image classification and can be primarily divided into score-based attacks and decision-based attacks~\cite{brendel2017decision,dong2020benchmarking}.

Under the score-based setting, attackers have access to the confidence score of the prediction, which can be used to guide the attack process~\cite{chen2017zoo, nesterov2017random}.
Most score-based attacks usually estimate the gradient by zero-order optimization methods through query access to the output of the target model~\cite{cheng2019improving}. 
Specifically, a perturbation is firstly initialized and added to the input image. The output score will guide the algorithm to find out the optimization direction of the next step.
For instance, Zeroth Order Optimization Based Black-box Attack (Zoo)~\cite{chen2017zoo} estimates the gradient at each coordinate by using the symmetric difference quotient.  
To improve the query efficiency, a random gradient-free (RGF) method~\cite{nesterov2017random} is proposed to get an approximated gradient by sampling random vectors independently from a distribution.

Different from score-based ones, attacks under decision-based setting are more challenging since only the final decision is provided for indicating whether the attacks succeed.
Existing decision-based attacks include Boundary Attack~(BA) ~\cite{brendel2017decision}, HopSkipJumpAttack (HSJA)~\cite{chen2020hopskipjumpattack}, {\it etc.}
They usually treat an irrelevant or target image as the start point and decrease the perturbation gradually to make the adversarial similar to input image visually~\cite{brendel2017decision, chen2020hopskipjumpattack, cheng2019sign}. 
However, most of these attacks are delicately proposed for the image classification tasks, and to the best of our knowledge, these still exists no query-based attacks for image retrieval.

%% file: latex/method.tex


In this section, we first formulate the problem of attacking image retrieval models under the black-box setting and then elaborate our proposed attack.
The whole attack pipeline is shown in Alg.~\ref{alg:attack},
given an input image $x$, we first conduct a white-box attack on a substitute model $s$ which is acquired with a recursive model stealing method beforehand (shown in Fig.~\ref{fig:model_stealing}), to provide the transfer-based priors for the following query-based attack.
Then, we quantify the attack effects
with a delicately designed relevance-based loss, 
and do gradient estimation following the basic idea of the score-based methods, aiming to provide the proper direction for the attack.
Finally, we repeat the aforementioned steps till the generated adversarial image $\hat{x}$ can deceive the target model successfully.

\subsection{Problem Formulation}
\label{sec:problem_formulation}

As shown in Fig.~\ref{fig:diff_cls_ret_2}, given a query image $x$, the image retrieval system with metric learning model $f$ and gallery \textbf{G} returns a list of images 
\begin{equation}
    \small{ \texttt{RList}}^n (x, f) = \{x_1, x_2, ..., x_i, ..., x_n | x_i\in \textbf{\text{G}} \},
\end{equation}
ordered by their similarities to $x$, where $n$ is the number of returned images
and $f$ projects $x$ to the feature space as $f(x)$.
In other words, $\mathcal{D}_f(x, x_i) \leq \mathcal{D}_f(x, x_j), s.t. \, i<j$ where $\mathcal{D}_f(x, x_i)=\lVert f(x) - f(x_i) \lVert^2_2$ is the metric that measures the feature distance between two images.

In this paper, the adversary aims to fool the target model into outputting a list of images whose top-$k$ has no overlap with original outputs under the assumption that the target model behaves well, \textit{i.e.}, the returned images are well organized according to the similarities to the input image.
Then, the attack goal can be formalized as
\begin{equation}
\small{\texttt{RList}}^{\mathcal{K}}(x, f) \cap 
\small{\texttt{RList}}^{\mathcal{K}}(x+\delta, f) = \O  \quad s.t. \quad {||\delta||}_p \leq \epsilon ,
\label{eq:def}
\end{equation}
where $\mathcal{K}$ is the number of top-ranked images to be considered and $\epsilon$ is the perturbation budget. 
$p$ determines the kind of tensor norm ($\ell_\infty$ by default) to measure the perturbation. 

The above goal can be solved by borrowing the idea of decision-based attacks proposed in the image classification task~\cite{chen2020hopskipjumpattack}, in which only the final decision (\textit{i.e.}, the predicted top label) instead of class probabilities is available to attackers.
However, as shown in Fig.~\ref{fig:landscape}(left), the loss landscape is discontinuous, it hence requires combinatorial optimization or exhaustive search algorithms, which may need a tremendous number of queries to perform a successful attack~\cite{cheng2018query}.

\begin{algorithm}[t!]
  \caption{\label{alg:attack} The query-based attack for image retrieval}
  \begin{algorithmic}[1]
    \Require{Target model $f$; input image $x$;  stolen model $s$; number of iterations for momentum $N_i$; max number of queries $T$; max perturbation $\epsilon$; step size $\sigma$; learning rate $\alpha$; number of considered images $\mathcal{K}$;}
    \State Initialize $\hat{x} \gets x, \mathcal{L}^{\text{prev}} \gets 1.0, y \gets \small{\texttt{RList}}^{\mathcal{K}}(x, f)$ 
    \For{$t \gets 1$ to $T/2$}
    \State \textit{\{Calculate basis $u$ with stolen model $s$\}} \Comment{Eq.~\ref{eq:MIM}}
    \State Initialize $\hat{x}^t  \gets \hat{x}, u \gets 0$
    \For{$i \gets 1 $ to $N_i$}
    \State $u = \beta \cdot u + \nabla_{\hat{x}^t}(\mathcal{L}_\text{w}(\hat{x}^{t}, y))$
    \State $\hat{x}^t = \textsc{clip}_{x,\epsilon}(\hat{x}^t + \alpha \cdot \text{sign}(u))$
    \EndFor
    \State \textit{\{Query attack with the resulted basis $u$\}} \Comment{Eq.~\ref{eq:RGF}}
    \State $\hat{g} \gets \frac{\mathcal{L}(\hat{x}+\sigma u, y) - \mathcal{L}(\hat{x}, y)}{\sigma} \cdot u$ 
    \State $\hat{x} \gets \textsc{clip}_{x,\epsilon}(\hat{x} + \alpha \cdot {\rm{sign}}(\hat{g}))$
    \If {$\mathcal{L}(\hat{x}, y) == \mathcal{L}^{\text{prev}}$}
        \State $\sigma \gets 2 \cdot \sigma$
    \EndIf
    \State  $\mathcal{L}^{\text{prev}} \gets \mathcal{L}(\hat{x}, y)$
    \EndFor \\
\Return adversarial sample $\hat{x}$
  \end{algorithmic}
\end{algorithm}

\subsection{Objective Function}
\label{sec:loss}
To solve the above problems, we delicately design an objective function to quantify the attack effects on the retrieval model to guide the generation of adversarial images.
Concretely, denote by $P(\hat{x}, y)$ the probability that the adversarial image $\hat{x}$ generated from the input $x$ fails to trigger the target model $f$ to misbehave, and denote by $y$ the true label of $x$, {\it i.e.}, $y=\small{\texttt{RList}}^{\mathcal{K}}(x, f)$.
Then, the objective is
\begin{align}
    {\rm{min}} \,\, \mathcal{L}(\hat{x}, y) = P(\hat{x}, y), \,\, s.t. \,\, ||\delta||_p \leq \epsilon.
\end{align}

To make $P(\hat{x}, y)$ computable, density estimation methods such as kernel density estimators~\cite{parzen1962estimation} can be applied. 
Since the computation cost is directly related to the number of samples, we need to sample as few samples as possible but approximate the distribution of $x$ as accurately as possible. 
We leverage the nearest neighbor density estimation method to approximate $P(\hat{x}, y)$ based on the top-$\mathcal{K}$ nearest neighbors of $x$ obtained by querying the target model.
Then, $P(\hat{x}, y)$ can be approximately rewritten as
\begin{eqnarray}
\footnotesize
\label{eqn:entropy}
    P(\hat{x}, y) \approx \sum_{i=1}^\mathcal{K}{P(x_i) P(\hat{x}, y | x_i)} = \sum_{i=1}^\mathcal{K}{\omega_i \cdot \varphi_i},
\end{eqnarray}
where $\omega_i=P(x_i)$ denotes the prior sampling probability and $\varphi_i=P(\hat{x}, y|x_i)$ denotes the conditioned attack failure probability.

\begin{figure}[!t]
 \center{\includegraphics[width=1.0\linewidth]{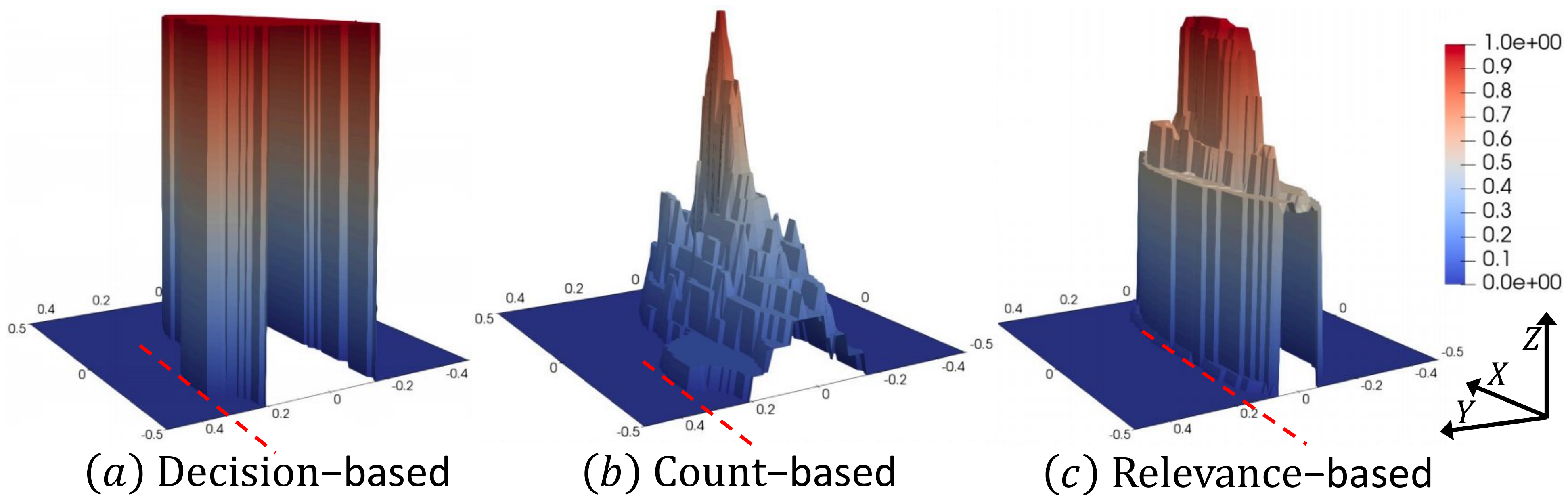}}
 \caption{\label{fig:landscape} Loss ($z$-axis) landscape visualization of the target retrieval model. Compared with perturbations with Gaussian ($x$-axis), the loss gets to 0 faster with adversarial perturbations ($y$-axis), showing the model's vulnerability against adversarial examples.
 We can find the hard-label problem is relaxed from left to middle. When relevance is considered, the loss gets to 0 with smaller perturbations.}
 \vspace{-3mm}
 \end{figure}

In the typical image retrieval system, $x_i$ is related to $x$ with a specific similarity score. However, this score cannot be obtained under the black-box setting.
The simplest strategy to tackle this problem is to treat each $x_i$ equally, \textit{i.e.}, $\forall i \in [1,\mathcal{K}], \omega_i = 1/\mathcal{K}$ (denoted as the \textbf{Count-based Loss}).
However, this is not the optimal strategy since it cannot reflect the attack effect in a fine-grained manner.
Recall that $x_i$ is ranked according to its similarity to $x$, thus we can use the ranking information to approximate their relevance.
Specifically, we refer to the Normalized Discounted Cumulative Gain metric~(NDCG) used in classical ranking
problem~\cite{jarvelin2017ir} and define $r_i$ as the relevance between $x_i$ and $x$. The probability $P(x_i)$ is defined as:
\begin{equation}
P(x_i) = \omega_i =
\frac{{2^{r_i}-1}}{\sum\nolimits_{i=1}^{\mathcal{K}}({2^{r_i}-1})}.
\label{eq:omega}
\end{equation}
With $r_i=\mathcal{K}-i$, the probability of the $i$-th result to be sampled is
a decaying exponential.


$\varphi_i = P(\hat{x}, y|x_i)$ indicates the attack failure probability of $\hat{x}$ given $x_i$. It can be obtained from the retrieved results. 
If $x_i \in \small{\texttt{RList}}^{\mathcal{K}}(\hat{x}, f)$, both $x$ and $\hat{x}$ are similar to $x_i$ and thus $\hat{x}$ should be similar to $x$, which also means the attack fails.
Considering the aforementioned rank-sensitive relevance and supposing that $x_i$ ranks at the $j$-th position in $\small{\texttt{RList}}^{\mathcal{K}}(\hat{x}, f)$, $\varphi_i$ can be denoted as 
\begin{equation}
\small
  \varphi_i = \begin{cases}
      \omega_j, \quad x_i \in \small{\texttt{RList}}^{\mathcal{K}}(\hat{x}, f) \,\,{\rm{and}}\,\, x_i=\hat{x}_j \\
      0, \quad x_i \not\in \small{\texttt{RList}}^{\mathcal{K}}(\hat{x}, f)
    \end{cases} .
\end{equation}

Then, the \textbf{Relevance-based} objective function $\mathcal{L}$ is rewritten as
 \begin{equation} \label{eq:loss}
     \mathcal{L}(\hat{x}, y) = \sum\nolimits_{i=1}^{\mathcal{K}}\omega_i \cdot
     {\varphi_i}, \,\, s.t. \,\, \|\delta\|_p \leq \epsilon.
 \end{equation}

In this way, the attack effects can be evaluated only based on $\small{\texttt{RList}}^{\mathcal{K}}(x, f)$ and $\small{\texttt{RList}}^{\mathcal{K}}(\hat{x}, f)$ according to Eq.~\ref{eq:loss}.
As shown in Fig.~\ref{fig:landscape}, compared to the count-based loss, the attack with relevance-based loss requires a smaller perturbation to reduce the loss to 0.


\subsection{Recursive Model Stealing}
\label{sec:model_stealing}
In adversarial attacks, the gradients for guiding the attack process are usually calculated based on the knowledge of the target model, which is unavailable under the black-box setting.
Thus, some studies try to use surrogate models to obtain prior-guided gradients and improve the attack efficiency~\cite{brunner2019guessing, cheng2019improving,guo2019subspace}.
However, 
the training data of the target model required for training a surrogate model is usually unavailable.
To tackle this problem, we propose to steal the gallery data of the IR system recursively via query access.

\begin{figure}[!t]        
  \center{\includegraphics[width=1.0\linewidth]{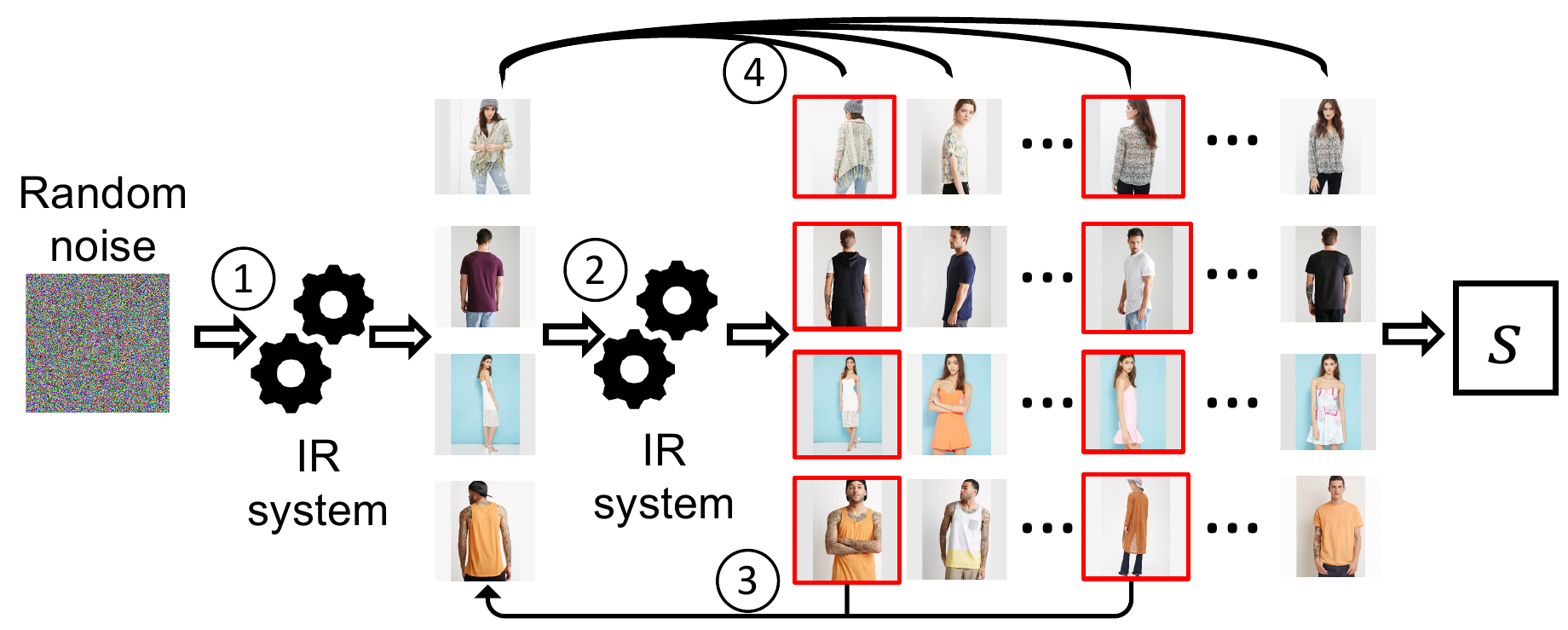}}
 \caption{\label{fig:model_stealing} The pipeline of our model stealing. First, an arbitrary image is put into the target image retrieval (IR) system (\textcircled{1}). The retrieved images are then evenly selected to construct a new query set, which will be put into the IR system in the next iteration ($\textcircled{2}, \textcircled{3}$) for more triplets. Finally, The stolen triplets ($\textcircled{4}$) will be used to train the substitute model $s$.}
 \vspace{-3mm}
 \end{figure}

Specifically, as shown in Fig.~\ref{fig:model_stealing}, queried by a random image $x$, the image retrieval system returns a set of retrieved images $\small{\texttt{RList}}^{n}(x, f)$, from which we select $N_c$ images evenly for greater diversity. 
These images again form a new image set as new queries to find more data.
The above procedure will be repeated for $C$ times to guarantee the diversity of collected images.
To better obtain the priors for attacks, the surrogate model should be trained to have a similar ranking capacity as the target IR model.
Hence, we query the target model with the collected $M$ images to get final triplets as the ground-truth for training the surrogate model $s$, of which the objective function is defined as
\begin{equation}
\small
    \sum\nolimits_{j>i} {\left[ \mathcal{D}_s(x, x_i) - \mathcal{D}_s(x, x_j) + \lambda \right]}_{+}.
    \label{eq:mse}
\end{equation}

We set $n=1000$, $N_c=10$, $C=3$ and $\lambda=0.05$ for all experiments. Thus, we only need 1111~(summed by 1+10+100+1000) queries to steal a model. 
Besides, the stolen model is dependent exclusively on the target model.
In addition to the stolen model, the stealing cost is also shared by all the test samples.
For example, the average query cost for each one is only $1 \approx 1111/1000$ if the number of test samples is 1000.

Our model stealing method is featured with the advantage that 
it requires no data beforehand.
This is quite different from model distillation  algorithms~\cite{li2019universal} which is usually performed based on the same training data with the target model.
It also differs from generative model based methods~\cite{zhou2020dast} , in which the collected data are usually out-of-distribution from the galleries of the target model. 
Besides, the diversity of the generated samples may be limited due to the problem of mode collapse.
In contrast, by querying target models constantly, we can steal data from the gallery in a recursive manner and guarantee the performance.

\subsection{Attack Optimization with Priors}
Since the decision-based problem in Eq.~\ref{eq:def} is relaxed with the proposed relevance-based loss, 
most of the query-based attacks proposed in image classification tasks can be extended to the retrieval tasks.
We therefore adopt RGF-attack~\cite{nesterov2017random} as our base framework and define its loss by the proposed relevance-based loss for the extension to retrieval systems.
The attack process can be summarized into two parts, {\it i.e.} gradient estimation and perturbation optimization.  
Denote $u_i$ as the $i$-th sampled basis vector which is sampled for $q$ times and $\hat{g}$ as the final estimated gradient. Then, gradient estimation and perturbation optimization are accomplished as follows:
\begin{equation}
\small
\begin{split}
    \hat{g} = \frac{1}{q}\sum_{i=1}^{q}\hat{g}_i, \,
    \hat{g}_i &= \frac{\mathcal{L}(x+\sigma u_i, y) - \mathcal{L}(x, y)}{\sigma} \cdot u_i, \\
    \hat{x} &= \textsc{clip}_{x,\epsilon}(x + \alpha \cdot {\rm{sign}}(\hat{g})),
\end{split}
\label{eq:RGF}        
\end{equation}
where $\sigma$ is the parameter to control the sampling variance and $\alpha$ is the learning rate.
The $\textsc{clip}_{x,\epsilon}$ operation aims to make the perturbation bounded in the budget~\cite{shi2019curls}. Besides, the generated adversarial example is converted to integer before it is fed into the target model to ensure its validity.

\begin{figure*}[htb]        
 \center{\includegraphics[width=1.0\linewidth]{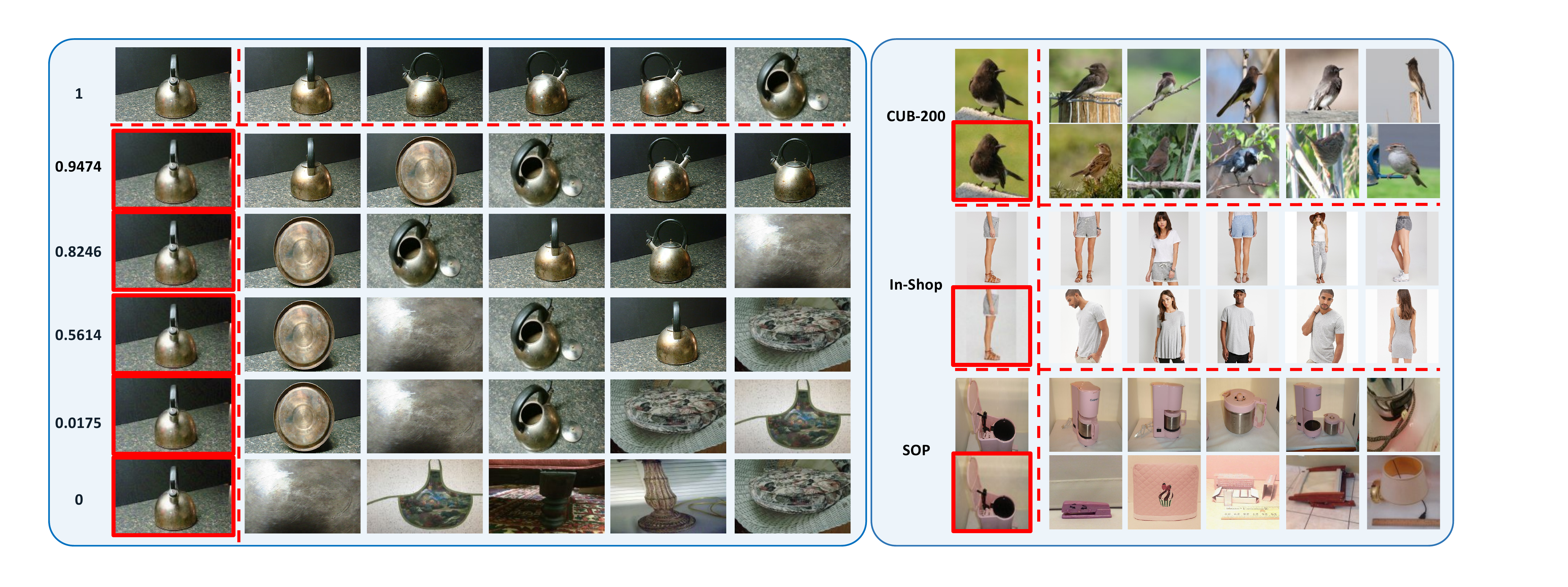}}
 \vspace{-3mm}
 \caption{\label{fig:result_procedure} Visualization of the attack procedures (left) and attack results (right). Images in red boxes are generated adversarial examples, which can fool the target model to return irrelevant images with imperceptible perturbations in the corresponding rows. Scores on the left are the corresponding loss to their searched results on their right. As more samples in the original sets disappear, the loss decays to 0.}
 \vspace{-4mm}
 \end{figure*}

\begin{table*}[h]
\renewcommand\arraystretch{1.2}
  \centering
  \footnotesize
 \setlength{\tabcolsep}{0.65mm}{
   \begin{tabular}{ c|cccccc|c|c|cccc|c|c|cccccc|c|c  }
   \toprule[1.2pt]
    \multirow{2}*{Attacks} &  \multicolumn{8}{c|}{CUB-200}  & \multicolumn{6}{c|}{SOP} & \multicolumn{8}{c}{In-Shop}  \\ \cline{2-23}
 	~ & 1  & 2 & 4 & 8 & 16 & 32 & AQ & ASR & 1 & 10 & 100 & 1000 & AQ & ASR & 1  & 10 & 20 & 30 & 40 & 50 & AQ & ASR\\
 	\hline
 	 	\hline
 	 Original & 
 	 0.61 & 0.73 & 0.87 & 0.91 & 0.98 & 0.99 & 0 & 0
 	 & 0.724 & 0.816 & 0.904 & 0.960 & 0 & 0
 	 & 0.642 & 0.868 & 0.910 & 0.926 & 0.938 & 0.945 & 0 & 0
 	  \\ \hline\hline
 	 \multicolumn{23}{c}{Comparsion with the State-of-the-art Methods}\\ \hline\hline
 	 OptAttack~\cite{cheng2018query} &
 	 \textbf{0.08} & \textbf{0.15} & \textbf{0.30} & 0.49 & 0.63 & 0.89 & 9708 & 0.04
 	 & 0.012 & 0.032 & 0.744 & 0.928 & 7931 & 0.288
 	 & \textbf{0.004} & \textbf{0.020} & 0.564 & 0.680 & 0.764 & 0.828 & 3017 & \textbf{0.948}
 	 \\
 	 
 	 Sign-Opt~\cite{cheng2019sign} &
 	 0.11 & 0.21 & 0.35 & 0.57 & 0.70 & 0.88 & 8833 & 0.00
 	 & 0.008 & 0.024 & 0.696 & 0.916 & 6746 & 0.372
 	 & 0.014 & 0.032 & 0.464 & 0.560 & 0.660 & 0.728 & 5564 & 0.492
 	 \\
 	 
 	 HSJA~\cite{chen2020hopskipjumpattack} & 
 	 0.13 & 0.23 & 0.41 & 0.47 & 0.60 & 0.92 & 10000 & 0.00
 	 & \textbf{0} & \textbf{0} & 0.632 & 0.880 & 5888 & 0.420
 	 & 0.004 & 0.024 & 0.316 & 0.470 & 0.564 & 0.624 & 5379 & 0.472\\
 	 \hline
 	 
 	 QAIR &
 	 0.16 & 0.23 & 0.32 & \textbf{0.45} & \textbf{0.56} & \textbf{0.76} & \textbf{93} & \textbf{0.69}
 	 & 0.016 & 0.064 & \textbf{0.472} & \textbf{0.832} & \textbf{35} & \textbf{0.904}
 	 & 0.008 & 0.044 & \textbf{0.132} & \textbf{0.256} & \textbf{0.312} & \textbf{0.352} & \textbf{35} & 0.916
 	 \\ \hline\hline
 	 
 	 \multicolumn{23}{c}{Component Analysis}\\ \hline\hline
 	 
 	 $\text{QAIR}_{C}$ &
 	 0.59 & 0.76 & 0.83 & 0.94 & 0.96 & 0.97 & 199 & 0.01
 	 & 0.176 & 0.372 & 0.724 & 0.916 & 113 & 0.480
 	 & 0.296 & 0.556 & 0.716 & 0.764 & 0.800 & 0.832 & 147 & 0.310
 	  \\ 
 	 $\text{QAIR}_{C-I}$ &
 	 0.36 & 0.51 &  0.60  &  0.74 & 0.80 & 0.92 & 152 & 0.32
 	 & 0.060 & 0.112 & 0.572 & 0.868 & 60 & 0.812
 	 & 0.072 & 0.164 & 0.272 & 0.396 & 0.476 & 0.520 & 65 & 0.784
 	  \\
 	 $\text{QAIR}_{C-S}$ &
 	 0.31 & 0.46 & 0.52 & 0.58 & 0.72 & 0.85 & 142 & 0.37
 	 & 0.056 & 0.088 & 0.532 & 0.848 & 51 & 0.836
 	 & 0.052 & 0.124 & 0.204 & 0.320 & 0.404 & 0.432 & 50 & 0.844
 	  \\ 
 	  
 	  \hline
 	  
   $\text{QAIR}_{R-S}$ &
 	 \textbf{0.16} & \textbf{0.23} & \textbf{0.32} & \textbf{0.45} & \textbf{0.56} & \textbf{0.76} & \textbf{93} & \textbf{0.69}
 	 & \textbf{0.016} & \textbf{0.064} & \textbf{0.472} & \textbf{0.832} & \textbf{35} & \textbf{0.904}
 	 & \textbf{0.008} & \textbf{0.044} & \textbf{0.132} & \textbf{0.256} & \textbf{0.312} & \textbf{0.352} & \textbf{35} & \textbf{0.916} \\
 	\hline
   \toprule[1.2pt]
   \end{tabular} 
   }
    \caption{
    Comparison with state-of-the-art methods on CUB-200, SOP and In-Shop before (Original) and after attack (others). Smaller Recall@K, smaller average number of queries (AQ) over both successful and failed attacks as well as higher Attack Success Rate (ASR) mean stronger attack.
    }
  \label{tbl:result}
  \vspace{-2mm}
\end{table*}

In RGF, the basis initialization is achieved by sampling random vectors independently from a distribution such as Gaussian. This can be improved with transferable priors~\cite{guo2019subspace}. 
For this, we follow the state-of-the art work Learnable Black-Box Attack~\cite{yang2020learning}, which utilizes the surrogate model $s$ to obtain the transfer-based priors and guide the basis selection.
Specifically, Momentum Iterative Method (MIM)~\cite{dong2018boosting} is firstly adopted to conduct the white-box attack based on the stolen model.
The derived momentum item $u$ is then used as the basis for the query-based attack. 
The loss $\mathcal{L}_\text{w}$ used for white-box attack is
\begin{equation}
\small
\mathcal{L}_\text{w}(\hat{x}, y) = ||s(\hat{x}) - \sum\nolimits_{i=1}^{\mathcal{K}} w_i \cdot s (x_i)||^2_2.
\label{eq:mse}
\end{equation}

And the optimization procedure for momentum is:
\begin{equation}
    \begin{split}
    u &= \beta \cdot u + \nabla_{\hat{x}}(\mathcal{L}_\text{w}(\hat{x}, y)), \\
    \hat{x} &= \textsc{clip}_{x,\epsilon}(\hat{x} + \alpha \cdot \text{sign}(u)),
    \end{split}
\label{eq:MIM}
\end{equation}
where $u$ is initialized with $\boldsymbol{0}$ and $\beta = 0.9$. The above procedure will be repeated for $N_i$ times.

Note that our QAIR is different from previous transfer-based attacks against image retrieval that utilize substitute models for crafting adversarial examples directly. Instead, QAIR employs the stolen model for obtaining transfer-based priors and generating prior-guided gradients for query attack. In this way, adversarial examples can be further rectified with query response until the attack succeeds.

%% file: latex/experiment.tex
In this section, we evaluate the proposed attack on various image retrieval models. 
More details and our demo video can be found in our supplementary material.

\begin{table*}[h]
  \centering
  \footnotesize
 \setlength{\tabcolsep}{1.5mm}{
   \begin{tabular}{ cc |cccccc |cccccc|c|c|c}
   \toprule[1.2pt]
    \multicolumn{2}{c|}{\multirow{2}{*}{Metric Learning Models}} & \multicolumn{6}{c|}{Recall@$K$ before our attacks} & \multicolumn{6}{c|}{Recall@$K$ after our attacks} & \multirow{2}{*}{AQ} & \multirow{2}{*}{ASR} & \multirow{2}{*}{DRR@1} \\ \cline{3-14}
 	 ~ & ~ & 1  & 2 & 4 & 8 & 16 & 32 & 1  & 2 & 4 & 8 & 16 & 32 & ~  & ~ & ~\\
 	\hline
 	 	\hline
 	 \multirow{4}{*}{BN-Inception~\cite{ioffe2015batch}} & Multi-Similarity~\cite{wang2019multi} & 0.61 & 0.73 & 0.87 & 0.91 & 0.98 & 0.99 
 	 & 0.16 & 0.23 & 0.32 & 0.45 & 0.56 & 0.76 & 93.40 & 0.69 & \textbf{73.77\%}\\ 
 	 
 	 ~ & Contrastive~\cite{hadsell2006dimensionality} & 0.57 & 0.66 & 0.81 & 0.89 & 0.92 & 0.96 
 	 & 0.16 & 0.28 & 0.45 & 0.55 & 0.64 & 0.78 & 89.99 & 0.68 & \textbf{71.93\%} \\
 	 
 	 ~ & HardMining~\cite{schroff2015facenet} & 0.62 & 0.75 & 0.81 & 0.88 & 0.94 & 0.97 
 	 & 0.24 & 0.29 & 0.39 & 0.48 & 0.62 & 0.75 & 96.34 & 0.64 & \textbf{61.29\%}\\
 	 
 	 ~ & Lifted~\cite{oh2016deep} & 0.62 & 0.73 & 0.84 & 0.92 & 0.94 & 0.97 
 	 &0.14 & 0.24 & 0.28 & 0.40 & 0.52 & 0.80 & 90.14 & 0.71 & \textbf{77.42\%}\\ \hline
 	 
 	 \multirow{4}{*}{DenseNet121~\cite{huang2017densely}} & Multi-Similarity~\cite{wang2019multi} & 0.66 & 0.81 & 0.89 & 0.94 & 0.96 & 0.99 
 	 & 0.08 & 0.15 & 0.24 & 0.37 & 0.53 & 0.67 & 84.46 & 0.72 & \textbf{87.88\%}\\
 	 
 	 ~ & Contrastive~\cite{hadsell2006dimensionality} & 0.66 & 0.80 & 0.88 & 0.91 & 0.95 & 0.98 
 	 & 0.10 & 0.15 & 0.23 & 0.29 & 0.42 & 0.60 & 83.80 & 0.70 & \textbf{84.85\%}\\
 	 
 	 ~ & HardMining~\cite{schroff2015facenet} & 0.66 & 0.76 & 0.85 & 0.92 & 0.97 & 0.99 
 	 &0.12 & 0.17 & 0.27 & 0.33 & 0.47 & 0.65 & 161.92 & 0.27 & \textbf{81.82\%}\\
 	 
 	 ~ & Lifted~\cite{oh2016deep} & 0.66 & 0.79 & 0.87 & 0.92 & 0.95 & 0.98 
 	 &0.07 & 0.15 & 0.26 & 0.41 & 0.51 & 0.65 & 84.36 & 0.68 & \textbf{89.39\%}\\
   \hline
   \toprule[1.2pt]
   \end{tabular} 
   }
    \caption{Recall@$K$ performances on the CUB-200 dataset before and after our attacks. It can be found that the proposed attack is effective on different image retrieval architectures trained with different metric learning methods. DRR@$1$ is the drop rate on Recall@$1$. The greater it is, the more vulnerable the image retrieval model is.}
  \label{tbl:BN_dense}
  \vspace{-3mm}
\end{table*}

\subsection{Experimental Settings}
\textbf{Datasets.} 
We evaluate our attack on three public datasets.
\noindent\textbf{Caltech-UCSD Birds-200-2011 (CUB-200)}~\cite{WahCUB_200_2011}: 
It has 200 classes of birds with 11788 images. The first 100 classes are split out for training and the rest for testing. It is a small but hard dataset for attack since it only has 100 classes in testing data.
\noindent\textbf{Stanford Online Products (SOP)}~\cite{oh2016deep}: 
It is a large scale dataset in image retrieval with $23k$ classes of $120k$ online product images from eBay.com.
It is split into 11,318 classes of 59,551 images for training and 11,316 classes of 60,502 images for testing.
\noindent\textbf{In-Shop Clothes (In-Shop)}~\cite{liuLQWTcvpr16DeepFashion}: This dataset contains 54642 images of 11735 clothing items from $Forever21$.
It provides 3,997 and 3,985 classes for training (25882 images) and testing (28760 images).

\textbf{Evaluation metrics.}
We use the commonly used metric Recall@$K$~\cite{oh2016deep} in image retrieval for evaluation. Greater drop of Recall@$K$ indicates stronger attack.
Besides, the commonly used attack success rate (ASR) metric in adversarial attack community is also employed. We treat the attack as successful when Eq.~\ref{eq:def} satisfies, thus ASR can be evaluated as the percentage of successful attacks. Note that ASR is designed for evaluating attacks against image retrieval under the black-box setting.
This is different from Recall@$K$ where true labels are required.

\textbf{Implementation details.}
We adopt the state-of-the-art image retrieval models$\footnote{https://github.com/bnu-wangxun/Deep\_Metric/\label{weight}}$~\cite{wang2019multi} as targets.
They are implemented with BN-Inception Network~\cite{ioffe2015batch} as most image retrieval works do for fairness and trained by their Multi-Similarity Loss.
The image retrieval results are listed in the Tab.~\ref{tbl:result} (The row with ``Original").
For model stealing, ResNet50~\cite{he2016deep} is adopted as the default backbone and trained with random horizontal flip and resized crop only since the pre-processing of the target model is unavailable to attackers. We evaluate on randomly sampled 250 images in the test sets on SOP and In-Shop (100 for CUB-200). The perturbation budget $\epsilon$ is set to 0.05 under $\ell_\infty$-norm and the max number of query $T$ is set to 200. 
The parameters in Eq.~\ref{eq:RGF} are set as follows: $q=1, \sigma=0.1, \alpha=0.01$~\cite{nesterov2017random}.
For each dataset, we set $\mathcal{K}=16$ in Eq.~\ref{eq:def} by default.

\subsection{Comparison with State-of-the-art Methods}
\begin{figure}[t]        
 \center{\includegraphics[width=1.0\linewidth, height=2.2in]{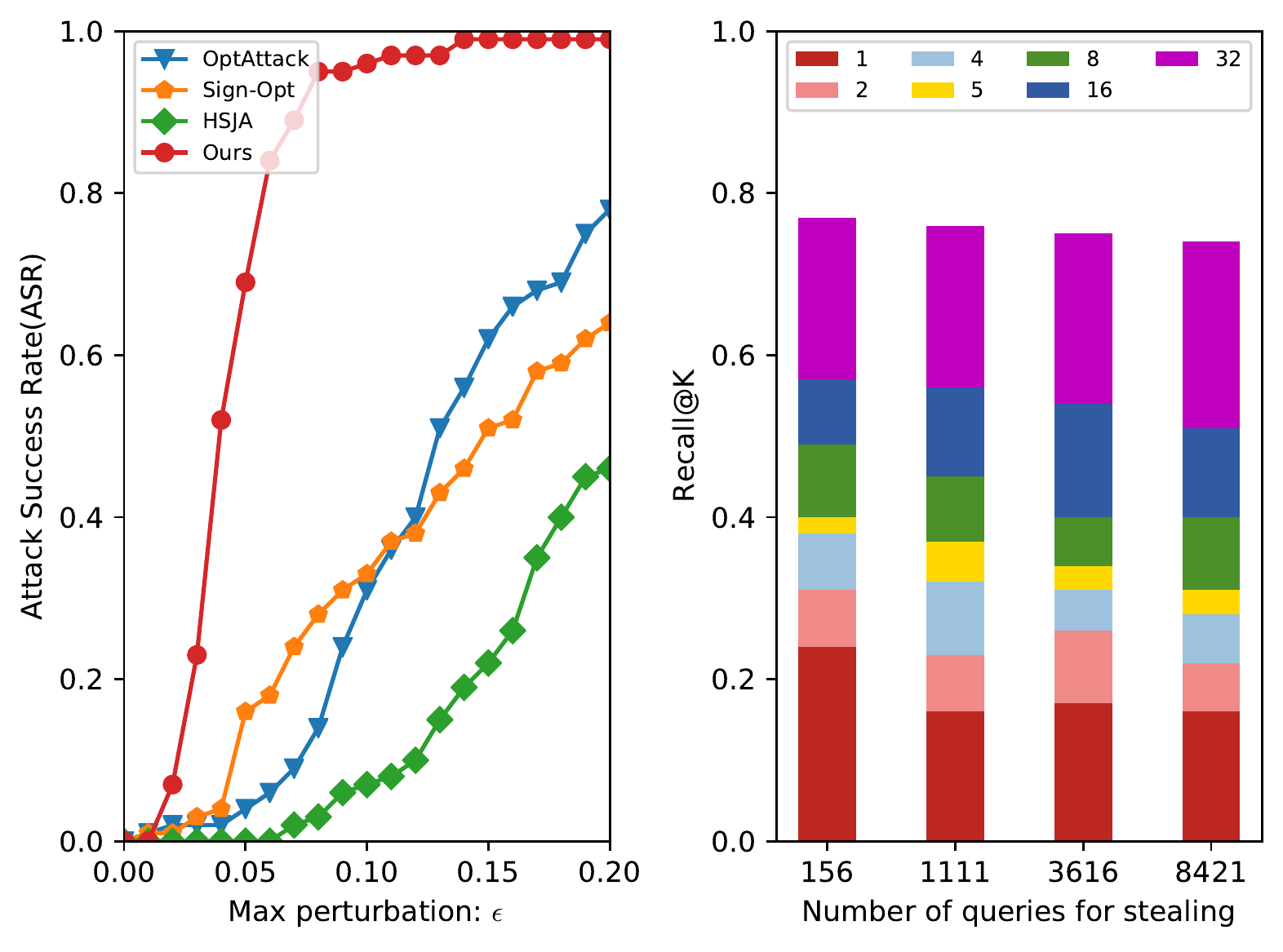}}
 \caption{\label{fig:query} Comparisons under different perturbation budgets on CUB-200 dataset (left) and 
 Recall@$K$ in terms of different numbers of queries used to steal the target model (right).}
 \vspace{-3mm}
 \end{figure}

Since the adversarial attack against image retrieval systems under black-box setting is a kind of decision-based attack, we compare our QAIR with several state-of-the-art decision-based attacks including Optimization-based attack (OptAttack)~\cite{cheng2018query}, Sign-Opt~\cite{cheng2019sign} and HopSkipJumpAttack (HSJA)~\cite{chen2020hopskipjumpattack}.
For these attacks, the maximum number of queries is set to 10000 to find adversarial examples with small perturbations.
As shown in Tab.~\ref{tbl:result}, our method can achieve comparable attack effects and at the same time, require much fewer queries. This proves the practical value of our method.
We found that though decision-based methods can completely subvert the top $\mathcal{K}$ results in most cases, the required maximum perturbation after 10000 queries is usually much higher than $\epsilon$, resulting in a low ASR. For a comprehensive study, we evaluate the ASR under different max perturbation limitations further. As shown in Fig.~\ref{fig:query} (left), our attack can always get a higher ASR than other methods, showing the effectiveness of the proposed approach. 
The visualization comparison of generated adversarial examples and comparison on defensive models can be found in our supplementary material.

\begin{table}[t]
  \centering
  \footnotesize
 \setlength{\tabcolsep}{2.mm}{
   \begin{tabular}{ cc|cccccc}
   \toprule[1.2pt]
    \multicolumn{2}{c|}{\multirow{2}{*}{Attacks}} & \multicolumn{6}{c}{CUB-200}\\ \cline{3-8}
 	 ~ & ~& 1  & 2 & 4 & 8 & 16 & 32 \\
 	\hline
 	 	\hline
 	 \multicolumn{2}{c|}{Original} & 0.61 & 0.73 & 0.87 & 0.91 & 0.98 & 0.99   \\ \hline
 	 \multirow{3}{*}{T} & FGSM~\cite{goodfellow2014explaining} & 0.33 & 0.45 & 0.56 & 0.66 & 0.76 & 0.85  \\
 	 ~ & BIM~\cite{kurakin2016adversarial} & 0.28 & 0.44 & 0.60 & 0.77 & 0.77 & 0.85  \\
 	 ~ & MIM~\cite{dong2018boosting} & 0.20 & 0.28 & 0.39 & 0.52 & 0.61 & \textbf{0.75}  \\ \hline
 	 Q & Ours (QAIR) & \textbf{0.16} & \textbf{0.23} & \textbf{0.32} & \textbf{0.45} & \textbf{0.56} & 0.76   \\
   \hline
   \toprule[1.2pt]
   \end{tabular} 
   }
    \caption{Comparison with transfer-based attacks (T). Q means query-based attack.}
  \label{tbl:tq}
  \vspace{-3mm}
\end{table}

\subsection{Comparison on Transfer and Query Attacks}
We also compare the proposed query-based attack with transfer-based attacks.
The evaluation results are listed in Tab.~\ref{tbl:tq}, from which we can see that the proposed attack outperforms transfer attacks developed based on different white-box attacks such as Fast Gradient Sign Method (FGSM)~\cite{goodfellow2014explaining} and Basic Iterative Method (BIM)~\cite{kurakin2016adversarial}, as well as the Momentum Iterative Method (MIM)~\cite{dong2018boosting}.
This is reasonable since the query-based attack can adjust the optimization direction with retrieval results, while transfer-based attack heavily relies on the transferability of generated adversarial examples.

\begin{table}[t]
  \centering
  \footnotesize
 \setlength{\tabcolsep}{1.3mm}{
   \begin{tabular}{ l |cccccc|c|c}
   \toprule[1.2pt]
    \multirow{2}*{Model} & \multicolumn{8}{c}{CUB-200}\\ \cline{2-9}
 	 & 1  & 2 & 4 & 8 & 16 & 32 & AQ  & ASR \\
 	\hline
 	 	\hline
 	 Original & 0.61 & 0.73 & 0.87 & 0.91 & 0.98 & 0.99 & 0 & 0 \\ \hline
 	 $S_{r18}$ & 0.18 & 0.24 & 0.35 & 0.53 & 0.62 & 0.78 & 92.14 & 0.70\\ 
 	 $S_{r50}$ & 0.16 & 0.23 & 0.32 & 0.45 & 0.56 & 0.76 & 93.40 & 0.69\\
 	 $S_{r101}$ & 0.24 & 0.29 & 0.35 & 0.44 & 0.60 & 0.79 & 99.92 & 0.65\\
 	 $S_{v16}$ & 0.28 & 0.37 & 0.44 & 0.52 & 0.65 & 0.82 & 121.84 & 0.54\\
 	 $S_{d121}$ & \textbf{0.14} & \textbf{0.23} & 0.34 & \textbf{0.42} & \textbf{0.55} & 0.77 & 87.64 & \textbf{0.73}\\
 	 $S_{d169}$ & 0.18 & 0.24 & \textbf{0.30} & 0.45 & 0.56 & \textbf{0.74} & \textbf{86.48} & 0.71\\
   \hline
   \toprule[1.2pt]
   \end{tabular} 
   }
    \caption{Recall@$K$ performance after our attack in terms of stolen models with different architectures.}
  \label{tbl:model}
\end{table}

\subsection{Attacks on Various Image Retrieval Models}
For comprehensive studies, we further evaluate the proposed methods on different target models trained with various metric learning methods (Multi-Similarity loss, Contrastive Loss~\cite{hadsell2006dimensionality}, Semi-Hard Mining Strategy~\cite{schroff2015facenet} and Lifted Structure Loss~\cite{oh2016deep}) on different architectures, including BN-Inception and Densenet121~\cite{huang2017densely}. 
As shown in Tab.~\ref{tbl:BN_dense}, the proposed attack can always produce a great Recall@$K$ drop against different image retrieval models, showing its generalization across models.

\begin{figure}[t]        
 \center{\includegraphics[width=1.0\linewidth, height=1.8in]{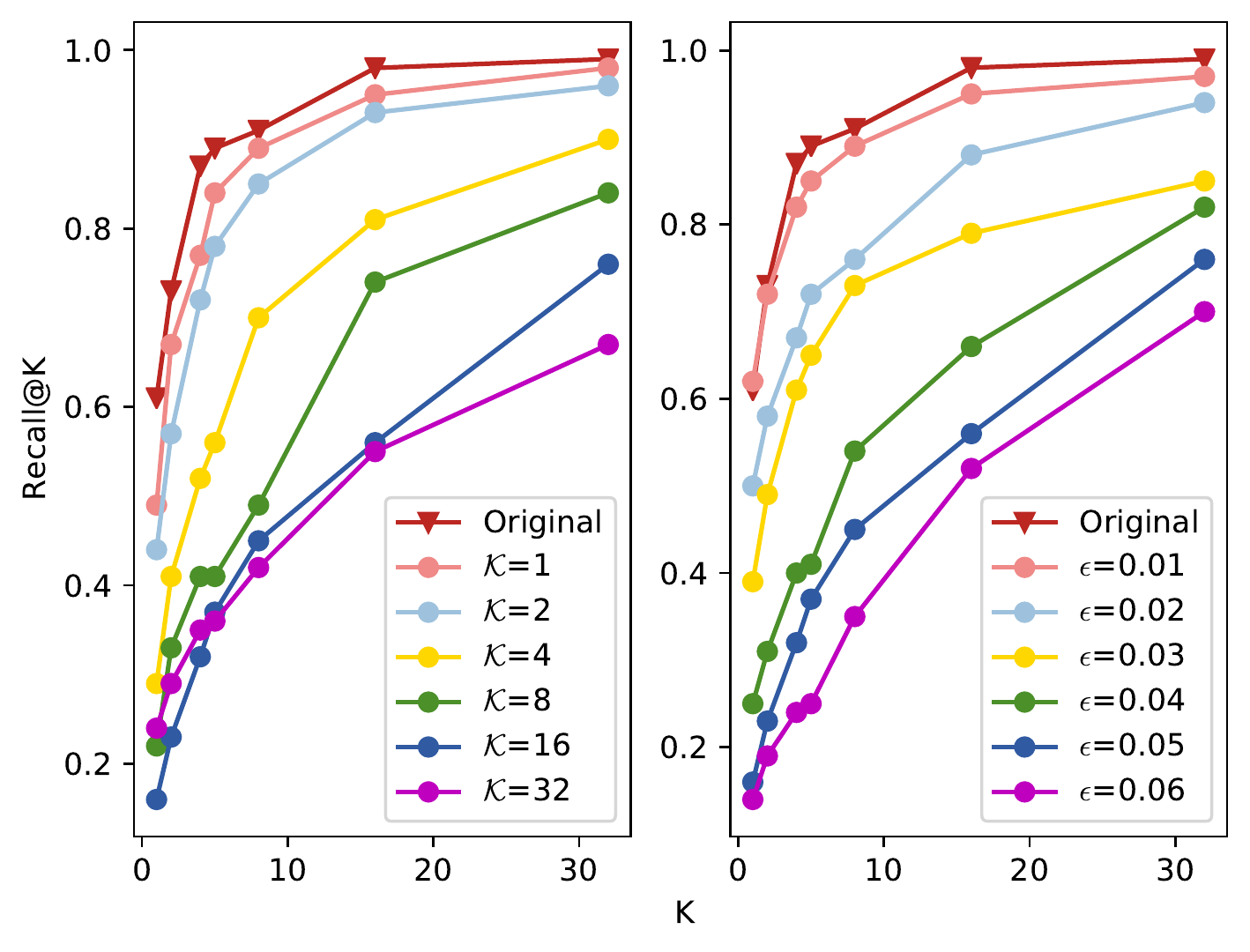}}
 \caption{\label{fig:loss} Recall@$K$ in terms of different number of considered candidates $\mathcal{K}$ (left) and perturbation budgets $\epsilon$ (right).}
 \vspace{-3mm}
 \end{figure}
 
\subsection{Ablation Study}
\textbf{Component analysis.}
    We treat RGF attack with the count-based loss as our baseline ($\text{QAIR}_{C}$), which samples bases from Gaussian distribution as RGF does.
    As shown in Tab.~\ref{tbl:result}, the proposed attack can already make a great drop on Recall@$K$ on both SOP and In-Shop datasets. This validates the vulnerability of large scale image retrieval models.
    To validate the effectiveness of the proposed model stealing, we compare attacks with bases provided by our stolen model  ($\text{QAIR}_{C-S}$) and model pretrained on ImageNet~\cite{russakovsky2015imagenet} ($\text{QAIR}_{C-I}$) for fairness. We can find that attacks with the stolen model are stronger under all datasets. Besides, the attack with relevance-based loss ($\text{QAIR}_{R-S}$) is much stronger than that with count-based loss, validating the effectiveness of our relevance-based loss. 

\textbf{Number of queries for model stealing.}
    We find that QAIR with bases from Gaussian distribution can already get a high ASR on large scale datasets, such as SOP and In-Shop, so we only evaluate the attack performances on CUB-200 dataset with different numbers of queries (156, 1111, 3616, 8421, corresponding to $N_c=5, 10, 15, 20$ respectively) for stealing model. As shown in Fig.~\ref{fig:query}, with more queries, the attack gets stronger. 

\textbf{Parameter analysis on $\mathcal{K}$ and $\epsilon$.}
    As shown in Fig.~\ref{fig:loss} (left), with larger $\mathcal{K}$, the attack leads to a greater Recall@$K$.
    With more candidates involved, the attacking procedure keeps more related samples of the original query  out of the attacked one.
    Recall@$K$ performances of attacks under different perturbation budgets can be found in Fig.~\ref{fig:loss} (right), from which we can see that a larger $\epsilon$ results in greater Recall@$K$ drop.

\textbf{Different model architectures for stolen model.}
    Tab.~\ref{tbl:model} shows the attack results with the stolen model from various model architectures. It can be found that the Recall@$K$ drops a lot under all circumstances, validating the universality of the proposed model stealing method.

\begin{figure}[t]        
 \center{\includegraphics[width=1.0\linewidth]{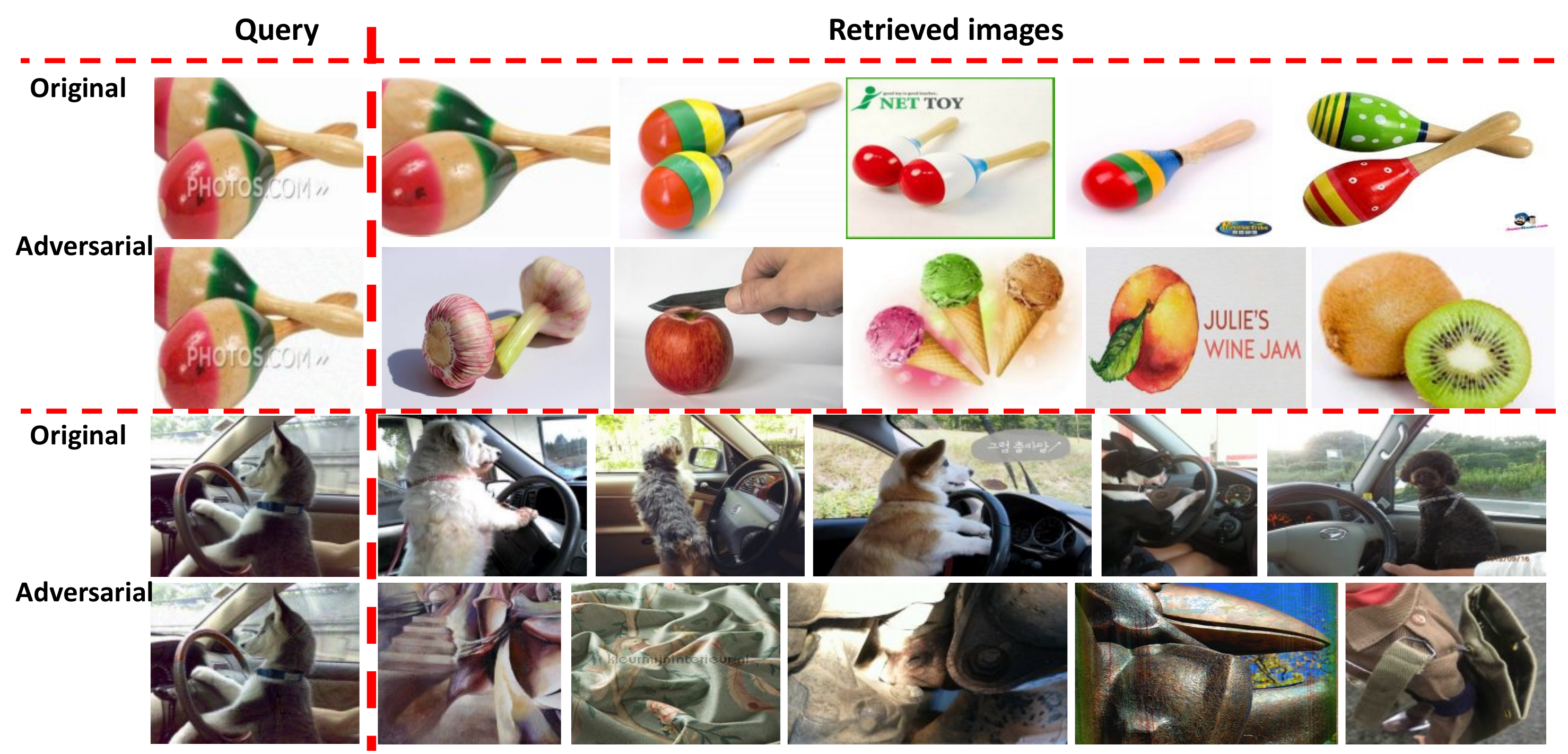}}
 \caption{\label{fig:bing} Searched results before (odd rows) and after (even rows) our attacks with Bing Visual Search.}
 \vspace{-5mm}
 \end{figure}

\subsection{Attacks on Real-world Commercial System}
Fig.~\ref{fig:bing} shows several attack examples generated against a real-world image retrieval system, {\it i.e.}, Bing Visual Search~\cite{hu2018web}. The even rows show the perturbed queries along with the retrieved images, which are completely different from the original ones at the odd rows. 
For the quantitative analysis,
we randomly sample 1000 images from ImageNet for testing and the proposed method can achieve 98\% attack success rate with only 33 queries on average, which demonstrates its practicability in real-world scenarios.

%% file: latex/conclusion.tex
In this paper, we introduce the first attempt on query-based attack against image retrieval under the black-box setting, where neither model parameter nor the training data is available. 
First, a relevance-based loss is designed to quantify the attack effects by measuring the set similarity on the top-$k$ retrieval results before and after the attack and guide the optimization of adversarial examples. To further boost the attack efficiency, a recursive model stealing method is proposed to obtain transfer-based priors and generate prior-guided gradients.
Extensive experiments show that the proposed attack achieves a high attack success rate with few queries against various image retrieval models.
Finally, the evaluation on the industrial visual search system further demonstrates the practical potential of the proposed method.
One limitation of our approach is that the attack may fail when the number of truly relevant images in the gallery is large, as shown in the supplementary material. In future work, we aim to go further for a more advanced objective towards stronger black-box attacks for developing robust image retrieval models.

%% file: latex/appendix.tex

\section{Appendices}
This supplementary material provides more details about the principles of loss landscapes (Fig. 2 in the paper) and decision-based attacks.
For comprehensive experiments, we also provide further evaluations on defensive models (Sec. 4.2 in the paper) and ablation studies on SOP and In-Shop datasets (Sec. 4.5 in the paper).
Besides, we also plot a scatter map to visualize the relationship between the Attack Success Rate (ASR) metric and Recall@$K$ drop for validating the rationality of the proposed attack goal experimentally.
More details about attacking real-world visual search engine are also provided (Sec. 4.6 in the paper).

\subsection{Loss Landscape}
The visualization of loss landscape is implemented with the toolbox provided by \cite{visualloss}. 
The loss is designed as follows:
\begin{equation}
    {\rm{loss}}(i, j) = \mathcal{L}(\hat{x}, y), \, s.t. \,\, \hat{x} = x + i * \gamma + j * \eta
\end{equation}
where coordinate $(i,j)$ determines the perturbation added on input image. $\gamma$ is a random direction sampled from Gaussian distribution while $\eta$ is the sign of gradient and can be generated with:
\begin{equation}
    \eta = {\rm{sign}}(u) = {\rm{sign}}(\frac{\partial (||f(\hat{x}) - f(x)||_2)}{\partial \hat{x}}).
\end{equation}
Note that the gradient is directly derived from the target model for its loss landscape visualization.

\begin{figure}[h]        
 \center{\includegraphics[width=0.9\linewidth]{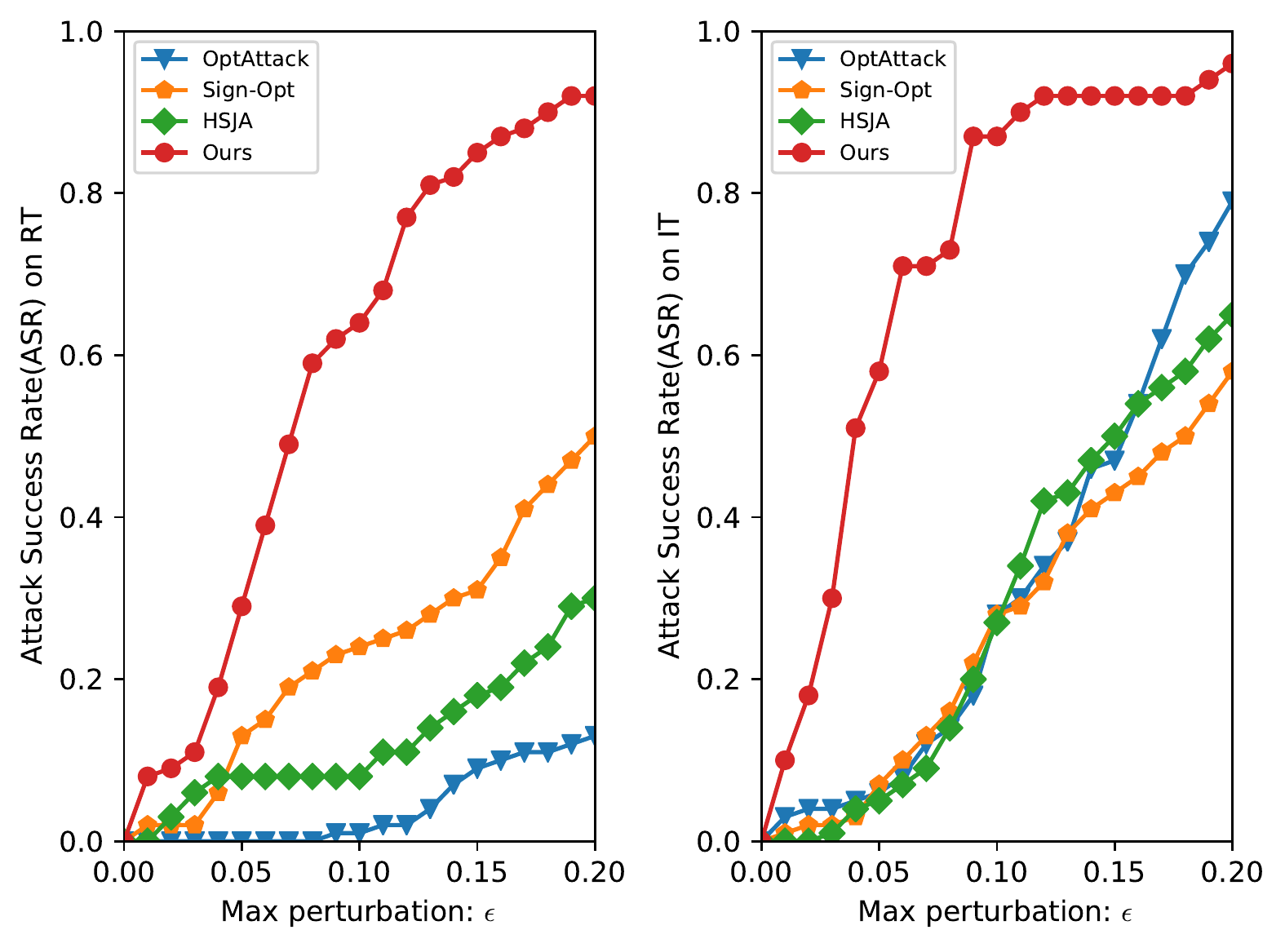}}
 \caption{\label{fig:defense} Attacks results on defensive models, including robust training~(RT, left) and input transformations (IT, right).}
 \end{figure}

\begin{figure}[h]        
 \center{\includegraphics[width=0.8\linewidth]{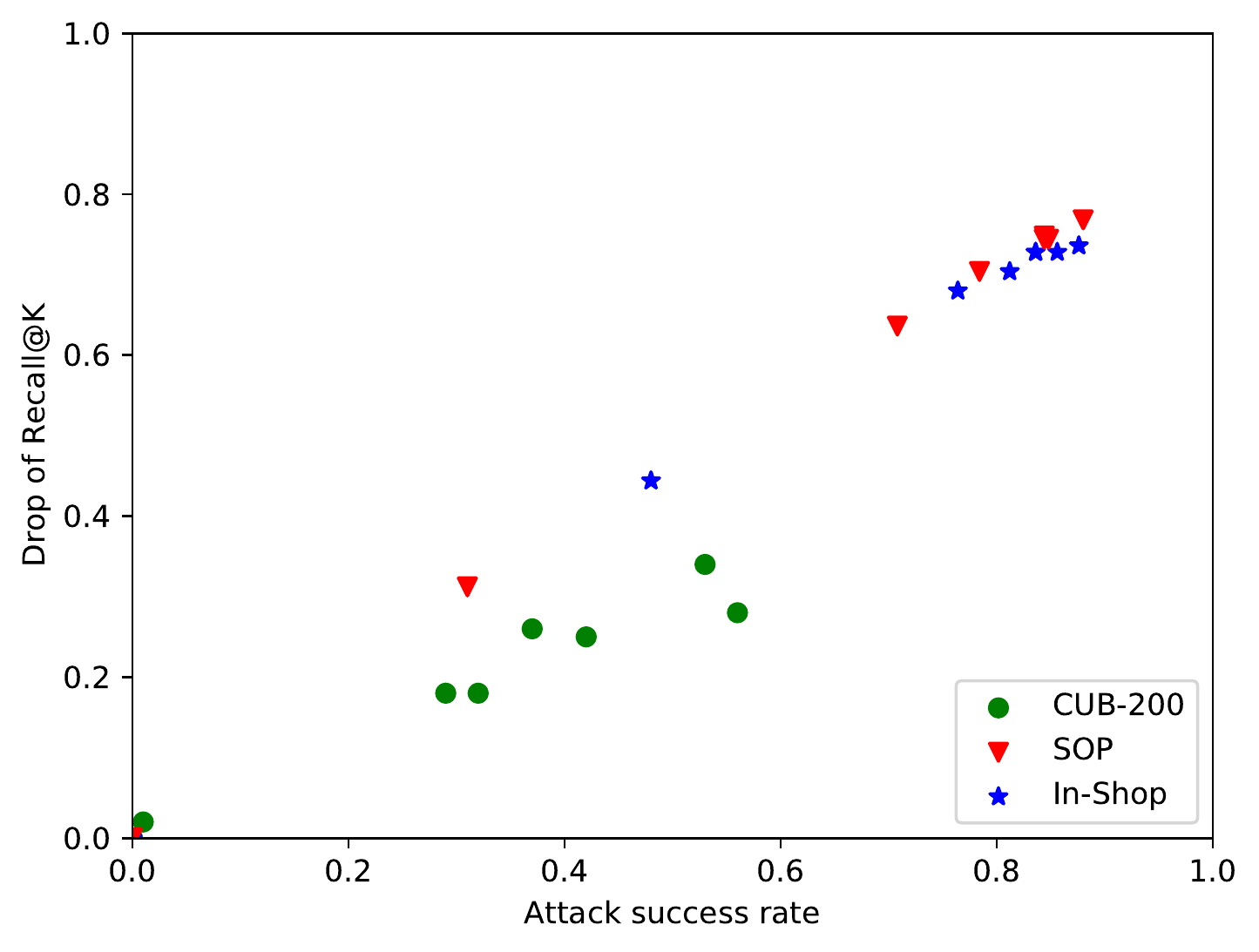}}
 \caption{\label{fig:ASR} Scatter map of ASR and drop of Recall@K metric.}
 \end{figure}

\begin{figure*}[t!]        
 \center{\includegraphics[width=0.95\linewidth]{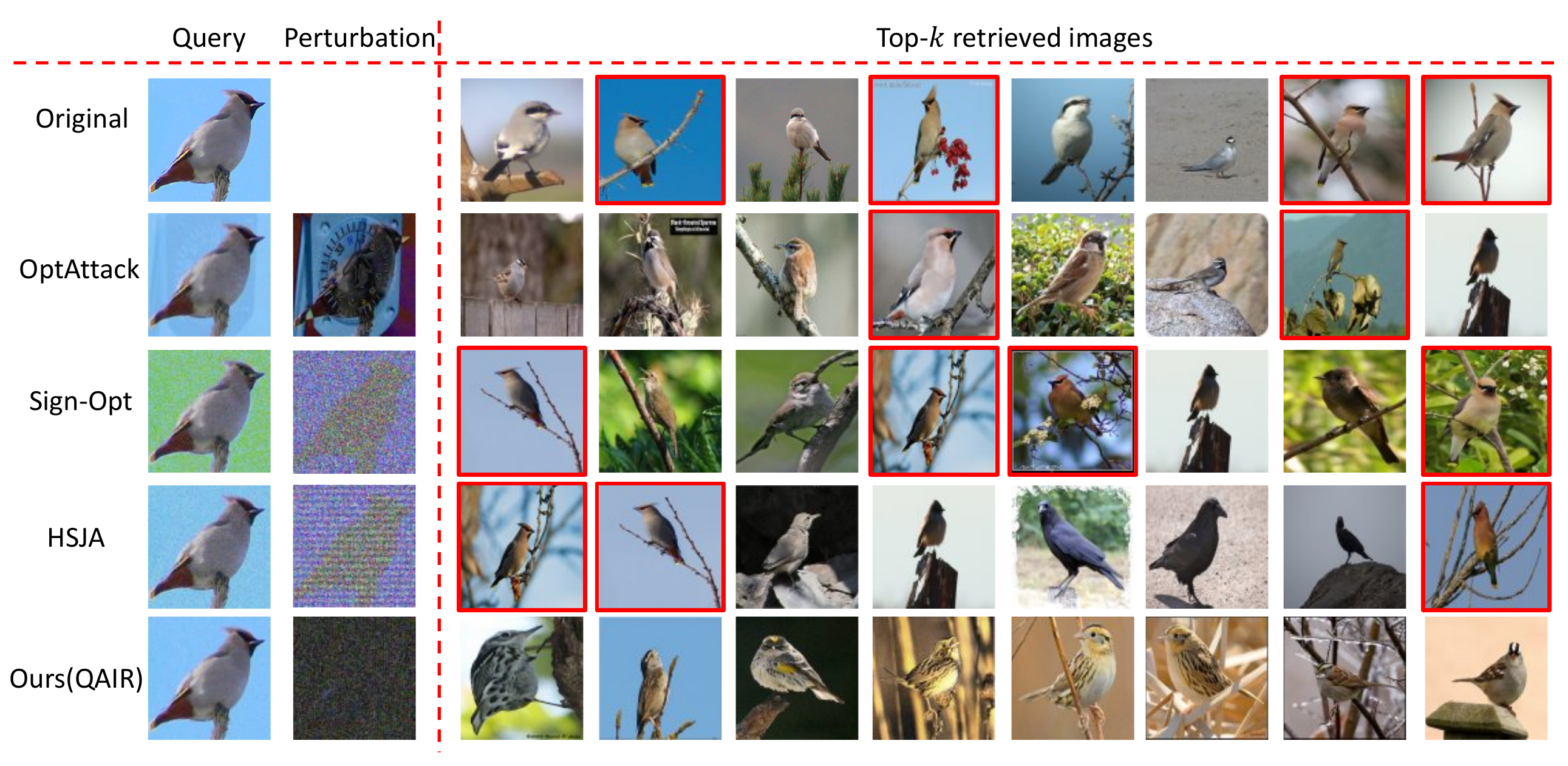}}
 \caption{\label{fig:compa} 
 Query results before and after attacks. Images in the first column are the queries. Images in the second column are the adversarial perturbations added on original images. Darker perturbation images mean smaller disturbances needed, which in turn indicates that the attacks tend to be more effective.
 The red boxes represent the correctly matched images.}
 \end{figure*}

\begin{table*}[!t]
  \centering
  \footnotesize
 \setlength{\tabcolsep}{1.0mm}{
   \begin{tabular}{ cc |cccccc |cccccc|c|c|c}
   \toprule[1.5pt]
    \multicolumn{2}{c|}{\multirow{2}{*}{Attacks}} & \multicolumn{6}{c|}{Recall@$K$ before attacks} & \multicolumn{6}{c|}{Recall@$K$ after attacks} & \multirow{2}{*}{AQ} & \multirow{2}{*}{ASR} & \multirow{2}{*}{DRR@1} \\ \cline{3-14}
 	 ~ & ~ & 1  & 10 & 20 & 30 & 40 & 50 & 1  & 10 & 20 & 30 & 40 & 50 & ~  & ~ & ~\\
 	\hline
 	 	\hline
 	 \multirow{4}{*}{BN-Inception~\cite{ioffe2015batch}} &
 	 Multi-Similarity~\cite{wang2019multi} & 0.853 & 0.959 & 0.965 & 0.973 & 0.976 & 0.979 
 	 & 0.008 & 0.044 & 0.132 & 0.256 & 0.312 & 0.352 & 35.19 & 0.92 & \textbf{99.06\%}
 	 \\ 
 	 
 	 ~ & Contrastive~\cite{hadsell2006dimensionality} & 0.832 & 0.956 & 0.976 & 0.980 & 0.984 & 0.984 
 	 & 0.008 & 0.068 & 0.124 & 0.260 & 0.320 & 0.372 & 38.93 & 0.90 & \textbf{99.04\%}
 	 \\
 	 
 	 ~ & HardMining~\cite{schroff2015facenet} & 0.868 & 0.980 & 0.988 & 0.992 & 0.996 & 0.996 
 	 & 0.028 & 0.112 & 0.208 & 0.336 & 0.412 & 0.464 & 57.32 & 0.82 & \textbf{96.77\%}
 	 \\
 	 
 	 ~ & Lifted~\cite{oh2016deep} & 0.828 & 0.944 & 0.960 & 0.972 & 0.976 & 0.988 
 	 & 0.032 & 0.080 & 0.172 & 0.292 & 0.380 & 0.436 & 42.51 & 0.88 & \textbf{96.14\%}
 	 \\ \hline
 	 
 	 \multirow{4}{*}{DenseNet121~\cite{huang2017densely}} & Multi-Similarity~\cite{wang2019multi} & 0.864 & 0.964 & 0.964 & 0.976 & 0.980 & 0.988 
    & 0.028 & 0.156 & 0.204 & 0.232 & 0.280 & 0.292 & 19.34 & 0.98 & \textbf{96.76\%}
 	 \\
 	 
 	 ~ & Contrastive~\cite{hadsell2006dimensionality} & 0.868 & 0.948 & 0.960 & 0.964 & 0.976 & 0.976 
 	 & 0.016 & 0.112 & 0.148 & 0.184 & 0.220 & 0.232 & 17.85 & 0.97 & \textbf{98.16\%}
 	 \\
 	 
 	 ~ & HardMining~\cite{schroff2015facenet} & 0.852 & 0.968 & 0.980 & 0.988 & 0.988 & 0.988 
 	 & 0.036 & 0.148 & 0.200 & 0.252 & 0.292 & 0.320 & 17.16 & 0.97 & \textbf{95.77\%}
 	 \\
 	 
 	 ~ & Lifted~\cite{oh2016deep} & 0.828 & 0.952 & 0.964 & 0.976 & 0.984 & 0.984 
 	 & 0.044 & 0.152 & 0.228 & 0.276 & 0.320 & 0.340 & 30.49 & 0.92 & \textbf{94.69\%}
 	 \\
   \hline
   \toprule[1.5pt]
   \end{tabular} 
   }
    \caption{Recall@$K$ performances on In-Shop dataset before and after attacks.}
  \label{tbl:BN_dense_shop_supp}
\end{table*}

\begin{table*}[t!]
  \centering
  \footnotesize
 \setlength{\tabcolsep}{1.0mm}{
   \begin{tabular}{ cc |cccc |cccc|c|c|c}
   \toprule[1.5pt]
    \multicolumn{2}{c|}{\multirow{2}{*}{Attacks}} & \multicolumn{4}{c|}{Recall@$K$ before attacks} & \multicolumn{4}{c|}{Recall@$K$ after attacks} & \multirow{2}{*}{AQ} & \multirow{2}{*}{ASR} & \multirow{2}{*}{DRR@1} \\ \cline{3-10}
 	 ~ & ~ & 1  & 10 & 100 & 1000 & 1  & 10 & 100 & 1000 & ~  & ~ & ~\\
 	\hline
 	 	\hline
 	 \multirow{4}{*}{BN-Inception~\cite{ioffe2015batch}} &
 	 Multi-Similarity~\cite{wang2019multi} & 0.729 & 0.855 & 0.932 & 0.978 
 	 & 0.016 & 0.064 & 0.472 & 0.832 & 35.45 & 0.90 & \textbf{97.81\%}
 	 \\ 
 	 
 	 ~ & Contrastive~\cite{hadsell2006dimensionality} & 0.701 & 0.839 & 0.920 & 0.975 
    & 0.008 & 0.028 & 0.440 & 0.792 & 27.57 & 0.94 & \textbf{98.86\%}
 	 \\
 	 
 	 ~ & HardMining~\cite{schroff2015facenet} & 0.723 & 0.861 & 0.937 & 0.980 
 	 & 0.016 & 0.032 & 0.428 & 0.824 & 31.73 & 0.94 & \textbf{97.79\%}
 	 \\
 	 
 	 ~ & Lifted~\cite{oh2016deep} & 0.703 & 0.839 & 0.923 & 0.974
 	 & 0.008 & 0.068 & 0.472 & 0.832 & 36.37 & 0.91 & \textbf{98.86\%}
 	 \\ \hline
 	 
 	 \multirow{4}{*}{DenseNet121~\cite{huang2017densely}} & Multi-Similarity~\cite{wang2019multi} & 0.720 & 0.824 & 0.904 & 0.964
 	 & 0.024 & 0.140 & 0.312 & 0.612 & 20.97 & 0.96 & \textbf{96.67\%}
 	 \\
 	 
 	 ~ & Contrastive~\cite{hadsell2006dimensionality} & 0.692 & 0.808 & 0.908 & 0.956
 	 & 0.040 & 0.136 & 0.356 & 0.660 & 19.62 & 0.96 & \textbf{94.22\%}
 	 \\
 	 
 	 ~ & HardMining~\cite{schroff2015facenet} & 0.706 & 0.842 & 0.927 & 0.976 
 	 & 0.048 & 0.144 & 0.340 & 0.620 & 19.10 & 0.97 & \textbf{93.20\%}
 	 \\
 	 
 	 ~ & Lifted~\cite{oh2016deep} & 0.704 & 0.808 & 0.900 & 0.964
 	 & 0.088 & 0.216 & 0.444 & 0.728 & 25.99 & 0.94 & \textbf{87.50\%}
 	 \\
   \hline
   \toprule[1.5pt]
   \end{tabular} 
   }
    \caption{Recall@$K$ performances on SOP dataset before and after attacks.}
  \label{tbl:BN_dense_sop_supp}
\end{table*}

\subsection{Decision-based Attack}
Decision-based attacks is a kind of query-based attack that requires only the decision of whether the attack succeeds. 
They usually treat an irrelevant or target image as a start point and decrease the perturbation gradually to make the adversarial similar to the input image visually during optimization~\cite{brendel2017decision, cheng2019sign, chen2020hopskipjumpattack}. 
For example, OptAttack~\cite{cheng2018query} starts the attack from an image that lies in the target class with a searched direction.
Then it reduces the distance of the perturbed image towards the original input in input space with binary search.
Though it can always succeed in subverting the outputs results in a great recall@$K$ drop, it requires a tremendous number of queries to achieve small perturbations. 

\subsection{Comparison on Defensive Models}
We further validate the effectiveness of the proposed method against several defensive models on CUB-200 dataset, including the classical robust training (RT)~\cite{goodfellow2014explaining} and input transformation (IT)~\cite{guo2017countering}. The results in Fig.~\ref{fig:defense} show that compared to state-of-the-art methods, our attack can achieve a much higher attack success rate under the same perturbation budgets.
This demonstrates the superiority of our method on attacking defensive models.

\subsection{Ablation Study on More Datasets}
Tab.~\ref{tbl:BN_dense_shop_supp} and Tab.~\ref{tbl:BN_dense_sop_supp} show more detailed experiments of attacking various deep metric learning models on In-Shop and SOP datasets, respectively. It can be found that the proposed query-based attack can achieve a high attack success rate on both datasets, demonstrating its effectiveness in different scenarios.

\subsection{Attack Goal and Objective Function}
Under the black-box setting, the attack success rate can only be calculated based on the observation of retrieved list. The rationality needs to be further explored. 
For this, we plot a scatter map of ASR and drop of Recall@$K$ (obtained based on the true label), which is shown in Fig.~\ref{fig:ASR} (under the same perturbations).
It can be found that the ASR is in proportion to Recall@$K$ drop, indicating the rationality of our attack goal experimentally. 
When comparing with state-of-the-art methods, the ASR metric takes both recall@$K$ drop and maximum perturbations into consideration, making it a relatively comprehensive metric.

For the proposed bidirectional relevance-based loss, we provide some examples to make it more comprehensible. Suppose only the top three candidates are considered. Given an input image $x$, the target image retrieval system will output the top three similar images $\{a_1, a_2, a_3\}$ and others $\{b_4, b_5, b_6, ...\}$. After attacks, there are several kinds of situations:
\begin{itemize}
\item $\{a_1, b_4, b_5\}$ and $\{a_3, b_4, b_5\}$. In general, higher rank denotes higher relevance to the input image $x$. Thus, the loss of situation $\{a_1, b_4, b_5\}$ should be greater than $\{a_3, b_4, b_5\}$. Thus, the rank-sensitive relevance before attacks should be considered.
\item $\{a_3, a_2, a_1\}$. The loss of situation $\{a_3, a_2, a_1\}$ should be smaller than $\{a_1, a_2, a_3\}$ since $a_1$ is the most relevant one to input image $x$. The lower it ranks, the more successful the attack is. Thus, how the candidates is ranked after attacks should also be considered.
\end{itemize}


\begin{figure}[h]        
 \center{\includegraphics[width=0.95\linewidth]{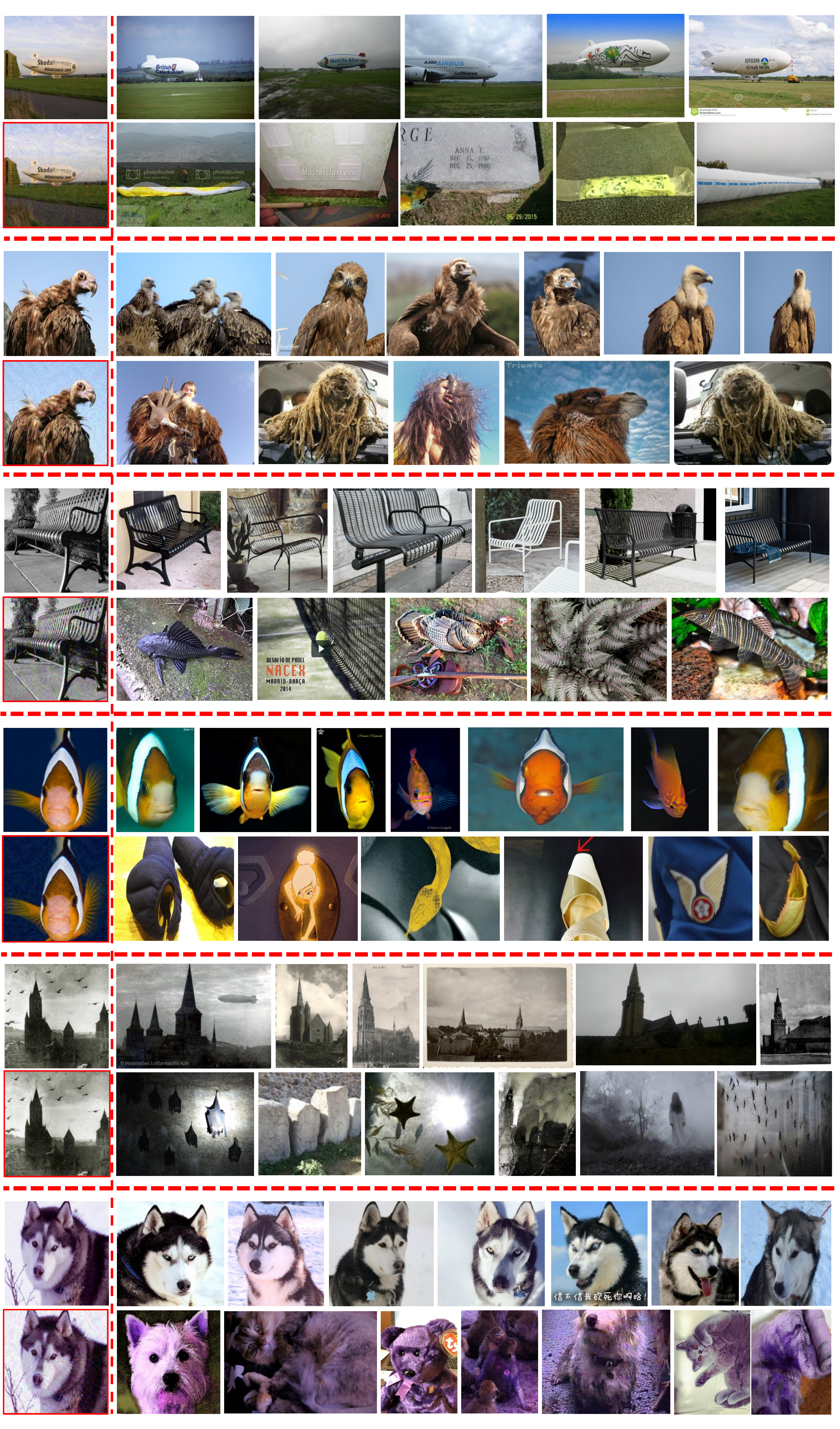}}
 \caption{\label{fig:bing_result} Query results on Bing Visual Search. Images in red boxes are adversarial examples generated with the proposed method. Images in the first column are queries while others are corresponding search results.}
 \end{figure}

\subsection{Visualization Comparison}
Fig.~\ref{fig:compa} shows the top 8 retrieved images of different input images~(the first column). Images in the red boxes are from the same category with the input query. It can be found that after 10,000 queries, the perturbations generated by other methods are still much greater than ours (darker perturbation images indicate smaller perturbations), which only needs 200 queries. 
Besides, though all the adversarial examples can subvert the top-$k$ retrieved results successfully, the retrieved results produced by other methods may contain images that share the same categories (red boxes) with the original ones. On the contrary, our method tries to push the adversarial example further from the original cluster in the feature space, which can relieve the inconsistency between attack goal and true labels.

\subsection{Attacks on Real-world Visual Search Engine}

\subsubsection{Implementation Details}
Unlike most existing transfer attacks, query-attack that we study in this paper needs to query search engine constantly. Bing Visual Search is the only image retrieval API that can be automatically queried. Thus, we only provided attack results on Bing Visual Search. 

Since Bing Visual Search is a frequently-used search engine and it has a huge amount of data in its gallery. Given an input image $x$, there are thousands of similar images with $x$. Thus, we need to take more candidates into consideration. 
For this, we set $\mathcal{K}=100$ to ensure the adversarial examples far away enough from the original clusters in the feature space.
Besides, the max query time and perturbation are set to 200 and 0.1. 
We only employ ResNet50 pretrained on ImageNet as our substitute model since it can make a good performance already. 

\subsubsection{Attack Results}
As shown in Fig.~\ref{fig:bing_result}, the generated adversarial examples can mislead the Bing Visual Search to output images actually irrelevant to the input image successfully with human-imperceptible perturbations. 
To quantitatively measure the performance, we randomly sample 1000 images from ImageNet for testing and the proposed method can achieve 98\% attack success rate with only 33 queries on average. This demonstrates the practicability of our attack in real-world scenarios.

We have also attached a video recording the image retrieval results on Bing Visual Search before and after attacks. It should be noted that Bing Visual Search updates their engine frequently. Thus some generated adversarial examples may be ineffective after a few days.

\begin{figure}[h]        
 \center{\includegraphics[width=1.0\linewidth]{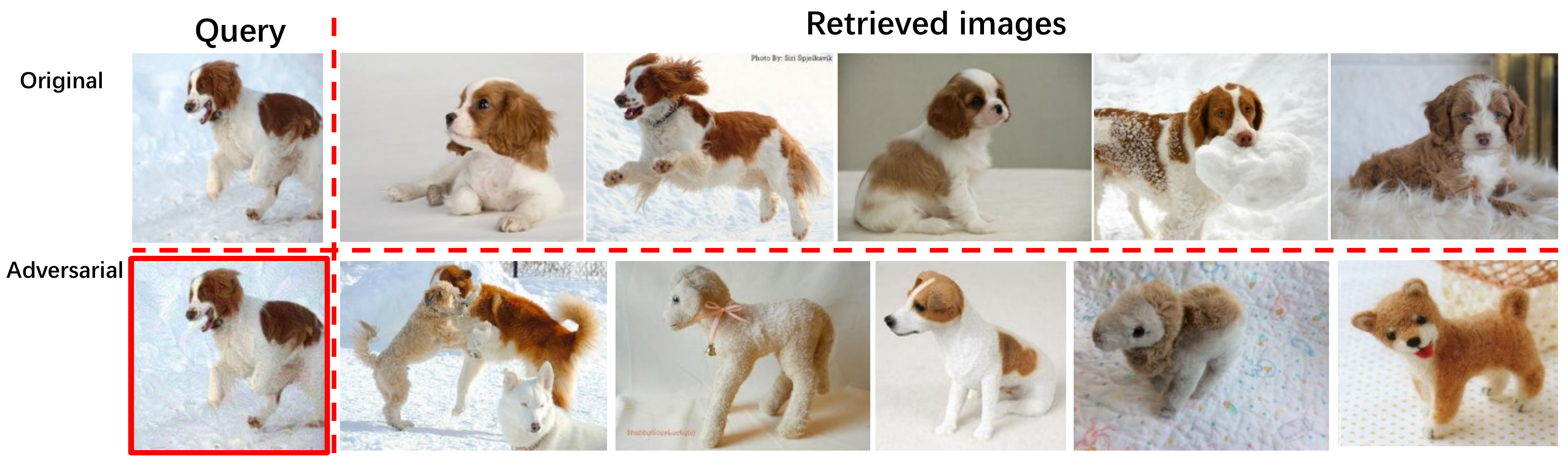}}
 \caption{\label{fig:fail} Failure attack examples. Image in red box is adversarial example generated with the proposed method. We can find that the retrieved images are still relevant to input image after attack even when its top-$k$ images are subverted.}
 \end{figure}
 
\begin{table}[h!]
  \small
  \centering
 \setlength{\tabcolsep}{1.2mm}{
 \begin{tabular}{c|c|cc|cc|cc}
 \toprule[1.5pt]
\multirow{2}{*}{Methods} & \multicolumn{1}{c|}{\multirow{2}{*}{MQ}} & \multicolumn{2}{c|}{CUB-200} & \multicolumn{2}{c|}{SOP}      & \multicolumn{2}{c}{In-Shop}  \\ \cline{3-8} 
~ &  ~ & ASR            & AQ          & ASR            & AQ          & ASR            & AQ          \\ \hline\hline
OptAttack~\cite{cheng2018query}  & \multirow{3}{*}{10,000}  & 0.04  & 9708 & 0.288 & 7931        & \textbf{0.948} & 3017        \\ 
Sign-Opt~\cite{cheng2019sign}  &  & 0              & 8833 & 0.372 & 6746  & 0.492 & 5564 \\ 
HSJA~\cite{chen2020hopskipjumpattack}                     &                                          & 0              & 10,000      & 0.420          & 5888        & 0.472          & 5379        \\ \hline
\multirow{3}{*}{Ours} & 200                                      & 0.69           & \textbf{93} & 0.904          & \textbf{35} & 0.916          & \textbf{35} \\ 
~ & 500                  & 0.72           & 180         & 0.918          & 64          & \textbf{0.924} & 58          \\ 
~                    & 1000                 & \textbf{0.73}  & 315         & \textbf{0.920} & 109         & 0.924          & 96 \\
\toprule[1.5pt]
\end{tabular}
}
\caption{Attack performance under different max query limitations (MQ). Higher attack success rate (ASR), smaller average queries (AQ) indicate stronger attacks.}
  \label{tbl:ASR_query}
\end{table} 

\subsection{Limitations and Future Work}
 
One limitation of the proposed method is that the attack may fail in practice even when the top-$k$ images are subverted, especially when the number of truly relevant images in the gallery is large, as shown in Fig.~\ref{fig:fail}. 
Apart from this, we also find the potential of the proposed QAIR may be limited due to the leverage of the substitute model. In OptAttack~\cite{cheng2018query} or traditional RGF attack~\cite{nesterov2017random}, they require lots of queries due to the randomness during optimization. Though we can improve the attack efficiency by leveraging the transfer-based priors as the guidance for optimization, the attack may fail due to the lack of adjustments of substitute model during attacks. Under this circumstance, more queries may not help much, as shown in Tab.~\ref{tbl:ASR_query}. In future work, we aim to go further for a more advanced objective and interactive model stealing method towards stronger black-box attacks for developing robust image retrieval models.

\bigskip
\noindent We would like to thank Bing Visual Search for making their API public and available.

%% file: cvpr.bbl
\begin{thebibliography}{10}\itemsep=-1pt

\bibitem{bai2019metric}
Song Bai, Yingwei Li, Yuyin Zhou, Qizhu Li, and Philip~HS Torr.
\newblock Metric attack and defense for person re-identification.
\newblock {\em arXiv preprint arXiv:1901.10650}, 2019.

\bibitem{bhagoji2018practical}
Arjun~Nitin Bhagoji, Warren He, Bo Li, and Dawn Song.
\newblock Practical black-box attacks on deep neural networks using efficient
  query mechanisms.
\newblock In {\em European Conference on Computer Vision}, pages 158--174.
  Springer, 2018.

\bibitem{brendel2017decision}
Wieland Brendel, Jonas Rauber, and Matthias Bethge.
\newblock Decision-based adversarial attacks: Reliable attacks against
  black-box machine learning models.
\newblock {\em arXiv preprint arXiv:1712.04248}, 2017.

\bibitem{brunner2019guessing}
Thomas Brunner, Frederik Diehl, Michael~Truong Le, and Alois Knoll.
\newblock Guessing smart: Biased sampling for efficient black-box adversarial
  attacks.
\newblock In {\em Proceedings of the IEEE International Conference on Computer
  Vision}, pages 4958--4966, 2019.

\bibitem{chen2020hopskipjumpattack}
Jianbo Chen, Michael~I Jordan, and Martin~J Wainwright.
\newblock Hopskipjumpattack: A query-efficient decision-based attack.
\newblock In {\em 2020 IEEE Symposium on Security and Privacy (SP)}, pages
  1277--1294. IEEE, 2020.

\bibitem{chen2017zoo}
Pin-Yu Chen, Huan Zhang, Yash Sharma, Jinfeng Yi, and Cho-Jui Hsieh.
\newblock Zoo: Zeroth order optimization based black-box attacks to deep neural
  networks without training substitute models.
\newblock In {\em Proceedings of the 10th ACM Workshop on Artificial
  Intelligence and Security}, pages 15--26, 2017.

\bibitem{chen2018shapeshifter}
Shang-Tse Chen, Cory Cornelius, Jason Martin, and Duen Horng~Polo Chau.
\newblock Shapeshifter: Robust physical adversarial attack on faster r-cnn
  object detector.
\newblock In {\em Joint European Conference on Machine Learning and Knowledge
  Discovery in Databases}, pages 52--68. Springer, 2018.

\bibitem{cheng2018query}
Minhao Cheng, Thong Le, Pin-Yu Chen, Jinfeng Yi, Huan Zhang, and Cho-Jui Hsieh.
\newblock Query-efficient hard-label black-box attack: An optimization-based
  approach.
\newblock {\em arXiv preprint arXiv:1807.04457}, 2018.

\bibitem{cheng2019sign}
Minhao Cheng, Simranjit Singh, Patrick Chen, Pin-Yu Chen, Sijia Liu, and
  Cho-Jui Hsieh.
\newblock Sign-opt: A query-efficient hard-label adversarial attack.
\newblock {\em arXiv preprint arXiv:1909.10773}, 2019.

\bibitem{cheng2019improving}
Shuyu Cheng, Yinpeng Dong, Tianyu Pang, Hang Su, and Jun Zhu.
\newblock Improving black-box adversarial attacks with a transfer-based prior.
\newblock In {\em Advances in Neural Information Processing Systems}, pages
  10934--10944, 2019.

\bibitem{dong2020benchmarking}
Yinpeng Dong, Qi-An Fu, Xiao Yang, Tianyu Pang, Hang Su, Zihao Xiao, and Jun
  Zhu.
\newblock Benchmarking adversarial robustness on image classification.
\newblock In {\em Proceedings of the IEEE/CVF Conference on Computer Vision and
  Pattern Recognition}, pages 321--331, 2020.

\bibitem{dong2018boosting}
Yinpeng Dong, Fangzhou Liao, Tianyu Pang, Hang Su, Jun Zhu, Xiaolin Hu, and
  Jianguo Li.
\newblock Boosting adversarial attacks with momentum.
\newblock In {\em Proceedings of the IEEE conference on computer vision and
  pattern recognition}, pages 9185--9193, 2018.

\bibitem{DPQN}
Yan Feng, Bin Chen, Tao Dai, and Shu-Tao Xia.
\newblock Adversarial attack on deep product quantization network for image
  retrieval.
\newblock {\em arXiv preprint arXiv:2002.11374v1}, 2020.

\bibitem{goodfellow2014explaining}
Ian~J Goodfellow, Jonathon Shlens, and Christian Szegedy.
\newblock Explaining and harnessing adversarial examples.
\newblock {\em arXiv preprint arXiv:1412.6572}, 2014.

\bibitem{guo2017countering}
Chuan Guo, Mayank Rana, Moustapha Cisse, and Laurens Van Der~Maaten.
\newblock Countering adversarial images using input transformations.
\newblock {\em arXiv preprint arXiv:1711.00117}, 2017.

\bibitem{guo2019subspace}
Yiwen Guo, Ziang Yan, and Changshui Zhang.
\newblock Subspace attack: Exploiting promising subspaces for query-efficient
  black-box attacks.
\newblock In {\em Advances in Neural Information Processing Systems}, pages
  3825--3834, 2019.

\bibitem{hadsell2006dimensionality}
Raia Hadsell, Sumit Chopra, and Yann LeCun.
\newblock Dimensionality reduction by learning an invariant mapping.
\newblock In {\em 2006 IEEE Computer Society Conference on Computer Vision and
  Pattern Recognition (CVPR'06)}, volume~2, pages 1735--1742. IEEE, 2006.

\bibitem{he2016deep}
Kaiming He, Xiangyu Zhang, Shaoqing Ren, and Jian Sun.
\newblock Deep residual learning for image recognition.
\newblock In {\em Proceedings of the IEEE conference on computer vision and
  pattern recognition}, pages 770--778, 2016.

\bibitem{hu2018web}
Houdong Hu, Yan Wang, Linjun Yang, Pavel Komlev, Li Huang, Xi Chen, Jiapei
  Huang, Ye Wu, Meenaz Merchant, and Arun Sacheti.
\newblock Web-scale responsive visual search at bing.
\newblock In {\em Proceedings of the 24th ACM SIGKDD international conference
  on knowledge discovery \& data mining}, pages 359--367, 2018.

\bibitem{huang2017densely}
Gao Huang, Zhuang Liu, Laurens Van Der~Maaten, and Kilian~Q Weinberger.
\newblock Densely connected convolutional networks.
\newblock In {\em Proceedings of the IEEE conference on computer vision and
  pattern recognition}, pages 4700--4708, 2017.

\bibitem{ioffe2015batch}
Sergey Ioffe and Christian Szegedy.
\newblock Batch normalization: Accelerating deep network training by reducing
  internal covariate shift.
\newblock {\em arXiv preprint arXiv:1502.03167}, 2015.

\bibitem{jarvelin2017ir}
Kalervo J{\"a}rvelin and Jaana Kek{\"a}l{\"a}inen.
\newblock Ir evaluation methods for retrieving highly relevant documents.
\newblock In {\em ACM SIGIR Forum}, volume~51, pages 243--250. ACM New York,
  NY, USA, 2017.

\bibitem{kariyappa2020maze}
Sanjay Kariyappa, Atul Prakash, and Moinuddin Qureshi.
\newblock Maze: Data-free model stealing attack using zeroth-order gradient
  estimation.
\newblock {\em arXiv preprint arXiv:2005.03161}, 2020.

\bibitem{kurakin2016adversarial}
Alexey Kurakin, Ian Goodfellow, and Samy Bengio.
\newblock Adversarial examples in the physical world.
\newblock {\em arXiv preprint arXiv:1607.02533}, 2016.

\bibitem{visualloss}
Hao Li, Zheng Xu, Gavin Taylor, Christoph Studer, and Tom Goldstein.
\newblock Visualizing the loss landscape of neural nets.
\newblock In {\em Neural Information Processing Systems}, 2018.

\bibitem{li2019universal}
Jie Li, Rongrong Ji, Hong Liu, Xiaopeng Hong, Yue Gao, and Qi Tian.
\newblock Universal perturbation attack against image retrieval.
\newblock In {\em Proceedings of the IEEE International Conference on Computer
  Vision}, pages 4899--4908, 2019.

\bibitem{liuLQWTcvpr16DeepFashion}
Ziwei Liu, Ping Luo, Shi Qiu, Xiaogang Wang, and Xiaoou Tang.
\newblock Deepfashion: Powering robust clothes recognition and retrieval with
  rich annotations.
\newblock In {\em Proceedings of IEEE Conference on Computer Vision and Pattern
  Recognition (CVPR)}, June 2016.

\bibitem{madry2017towards}
Aleksander Madry, Aleksandar Makelov, Ludwig Schmidt, Dimitris Tsipras, and
  Adrian Vladu.
\newblock Towards deep learning models resistant to adversarial attacks.
\newblock {\em arXiv preprint arXiv:1706.06083}, 2017.

\bibitem{nesterov2017random}
Yurii Nesterov and Vladimir Spokoiny.
\newblock Random gradient-free minimization of convex functions.
\newblock {\em Foundations of Computational Mathematics}, 17(2):527--566, 2017.

\bibitem{oh2016deep}
Hyun Oh~Song, Yu Xiang, Stefanie Jegelka, and Silvio Savarese.
\newblock Deep metric learning via lifted structured feature embedding.
\newblock In {\em Proceedings of the IEEE conference on computer vision and
  pattern recognition}, pages 4004--4012, 2016.

\bibitem{papernot2017practical}
Nicolas Papernot, Patrick McDaniel, Ian Goodfellow, Somesh Jha, Z~Berkay Celik,
  and Ananthram Swami.
\newblock Practical black-box attacks against machine learning.
\newblock In {\em Proceedings of the 2017 ACM on Asia conference on computer
  and communications security}, pages 506--519, 2017.

\bibitem{parzen1962estimation}
Emanuel Parzen.
\newblock On estimation of a probability density function and mode.
\newblock {\em The annals of mathematical statistics}, 33(3):1065--1076, 1962.

\bibitem{russakovsky2015imagenet}
Olga Russakovsky, Jia Deng, Hao Su, Jonathan Krause, Sanjeev Satheesh, Sean Ma,
  Zhiheng Huang, Andrej Karpathy, Aditya Khosla, Michael Bernstein, et~al.
\newblock Imagenet large scale visual recognition challenge.
\newblock {\em International journal of computer vision}, 115(3):211--252,
  2015.

\bibitem{schroff2015facenet}
Florian Schroff, Dmitry Kalenichenko, and James Philbin.
\newblock Facenet: A unified embedding for face recognition and clustering.
\newblock In {\em Proceedings of the IEEE conference on computer vision and
  pattern recognition}, pages 815--823, 2015.

\bibitem{shan2020fawkes}
Shawn Shan, Emily Wenger, Jiayun Zhang, Huiying Li, Haitao Zheng, and Ben~Y
  Zhao.
\newblock Fawkes: Protecting privacy against unauthorized deep learning models.
\newblock In {\em 29th $\{$USENIX$\}$ Security Symposium ($\{$USENIX$\}$
  Security 20)}, pages 1589--1604, 2020.

\bibitem{shi2019curls}
Yucheng Shi, Siyu Wang, and Yahong Han.
\newblock Curls \& whey: Boosting black-box adversarial attacks.
\newblock In {\em Proceedings of the IEEE Conference on Computer Vision and
  Pattern Recognition}, pages 6519--6527, 2019.

\bibitem{WahCUB_200_2011}
C. Wah, S. Branson, P. Welinder, P. Perona, and S. Belongie.
\newblock {The Caltech-UCSD Birds-200-2011 Dataset}.
\newblock Technical Report CNS-TR-2011-001, California Institute of Technology,
  2011.

\bibitem{wang2020transferable}
Hongjun Wang, Guangrun Wang, Ya Li, Dongyu Zhang, and Liang Lin.
\newblock Transferable, controllable, and inconspicuous adversarial attacks on
  person re-identification with deep mis-ranking.
\newblock In {\em Proceedings of the IEEE/CVF Conference on Computer Vision and
  Pattern Recognition}, pages 342--351, 2020.

\bibitem{wang2019multi}
Xun Wang, Xintong Han, Weilin Huang, Dengke Dong, and Matthew~R Scott.
\newblock Multi-similarity loss with general pair weighting for deep metric
  learning.
\newblock In {\em Proceedings of the IEEE Conference on Computer Vision and
  Pattern Recognition}, pages 5022--5030, 2019.

\bibitem{yang2020learning}
Jiancheng Yang, Yangzhou Jiang, Xiaoyang Huang, Bingbing Ni, and Chenglong
  Zhao.
\newblock Learning black-box attackers with transferable priors and query
  feedback.
\newblock 2020.

\bibitem{zhao2019unsupervised}
Guoping Zhao, Mingyu Zhang, Jiajun Liu, and Ji-Rong Wen.
\newblock Unsupervised adversarial attacks on deep feature-based retrieval with
  gan.
\newblock {\em arXiv preprint arXiv:1907.05793}, 2019.

\bibitem{zheng2018open}
Zhedong Zheng, Liang Zheng, Zhilan Hu, and Yi Yang.
\newblock Open set adversarial examples.
\newblock {\em arXiv preprint arXiv:1809.02681}, 2018.

\bibitem{zhou2020adversarial}
Mo Zhou, Zhenxing Niu, Le Wang, Qilin Zhang, and Gang Hua.
\newblock Adversarial ranking attack and defense.
\newblock {\em arXiv preprint arXiv:2002.11293}, 2020.

\bibitem{zhou2020dast}
Mingyi Zhou, Jing Wu, Yipeng Liu, Shuaicheng Liu, and Ce Zhu.
\newblock Dast: Data-free substitute training for adversarial attacks.
\newblock In {\em Proceedings of the IEEE/CVF Conference on Computer Vision and
  Pattern Recognition}, pages 234--243, 2020.

\end{thebibliography}
